\documentclass{article}

\usepackage{arxiv}
\usepackage[utf8]{inputenc}
\usepackage[T1]{fontenc}
\usepackage{url}
\usepackage{booktabs}
\usepackage{amsfonts}
\usepackage{nicefrac}
\usepackage{microtype}
\usepackage{graphicx}

\usepackage{doi}

\usepackage{amsmath}
\usepackage{array}

\usepackage{caption}
\usepackage{float}

\usepackage{hyperref}

\usepackage{listings}
\usepackage{xpatch}

\usepackage{algorithm}
\usepackage{algpseudocode} 
\newcommand{\INDSTATE}[1]{\State\hspace{#1}}

\usepackage[table]{xcolor}

\usepackage[numbers,sort&compress]{natbib}

\definecolor{lightgrey}{rgb}{0.95, 0.95, 0.95}
\definecolor{darkgrey}{rgb}{0.6, 0.6, 0.6}

\usepackage{tikz}

\newcolumntype{"}{@{\vrule width 1pt}} 
\newcolumntype{!}{@{}>{\kern2pt}} 

\title{EvoGPT-f: An Evolutionary GPT Framework for Benchmarking Formal Math Languages\\[1ex]}
\author{ \href{https://www.linkedin.com/in/johnmercer/}
        {Johnathan D.~Mercer} \\
	Boston, MA  \\
	\texttt{presci3nce@gmail.com} \\
}

\renewcommand{\shorttitle}{EvoGPT-f: Benchmarking Machine Learnability of Formal Mathematics Languages}

\hypersetup{
pdftitle={EvoGPT-f: An Evolutionary Framework for Benchmarking Formal Math Languages},
pdfsubject={},
pdfauthor={Johnathan~Mercer},
pdfkeywords={GPT, transformer, type theory, Lean, Coq, HOL, genetic algorithms, automated theorem proving, interactive theorem proving},
}

\begin{document}

\date{February 11, 2024}
\maketitle

\begin{abstract}
Formal mathematics is the discipline of translating mathematics into a programming language in which any statement can be unequivocally checked by a computer. Mathematicians and computer scientists have spent decades of painstaking formalization efforts developing languages such as Coq, HOL, and Lean. Machine learning research has converged on these formal math corpora and given rise to an assortment of methodologies to aid in interactive and automated theorem proving. However, these papers have primarily focused on one method, for one proof task, in one language. This paper introduces EvoGPT-f: a novel evolutionary framework for the first systematic quantitative analysis of the differential machine learnability of five formal math corpora (Lean 3, Lean 4, Coq, HOL 4, HOL Light) using four tokenization methods (character, word-level, Byte Pair Encoding and StarCoder tokenizer). This paper does not put to rest the question of the ``best" or ``easiest" language to learn. Rather, this framework and preliminary findings begin to illuminate the differential machine learnability of these languages, offering a foundation to forge more systematic quantitative and qualitative comparative research across communities.
\end{abstract}

\vspace{80mm}
\section{Introduction}

Decades of perseverance and collaboration between mathematicians and computer scientists have led to a diverse set of formal mathematics languages, proof checkers, and communities supporting their development. These include, but are not limited to: Lean\cite{mathlib2020}, Coq\cite{Coq2017}, HOL\cite{HOL1993} family (such as Isabelle \cite{Isabelle2002} and HOL Light \cite{HOLLightPaper}), Metamath\cite{Metamath2019}, and Mizar\cite{MizarWebsite}. Despite ideological and cultural differences (outlined e.g. in \cite{wiedijk_mizar}), these communities continue the painstaking work required to realize the promise of unequivocal mathematical proofs and large-scale mathematical databases \cite{CADE-1994-Anonymous}.
 
These formal mathematics codebases and corresponding proof checkers converged with the modern developments of machine learning and artificial intelligence to usher in vibrant research in the application of supervised machine learning and neural language models \citep{DeepMath2016, Holophrasm2016, DeepNetworkGuidedProofSearch2017, FormulaNet2017, rlCoP2018,GamePad2018, HOList2019, MetaGen2020, GPT-f2020, LIME2021, PACT2021, HTPS2022, yang2023leandojo}, as well as evolutionary methods \cite{TheoremProvingHeuristicsWithGA2015, ETheoremProver, DEAP_JMLR2012, EvoATP2016, EvoSearchWithHF2017, EvoTheoremProvingIsabelleHOL2019, EvoHeuristicProofSearchingHOL42021, EvoHeuristicSearchForITP2021} to aid in premise selection, proof guidance, and proof search in Interactive Theorem Proving(ITP) and Automated Theorem Proving(ATP). 

Previous research has been primarily focused on applying one method, for one proof task, using one formal language. This is, in part, due to the fact that historically the choice of formal language is often determined based on where you study. The Coq user base is driven by Inria in France, Lean by Microsoft in the US, and HOL by the University of Cambridge in the UK. Aside from nominal qualitative head-to-head comparisons of languages \cite{Zammit_HOL_vs_Coq}, there does not exist a systematic quantitative analysis on the differential machine learnability of these languages.

To motivate more systematic quantitative and qualitative analyses across formal languages, I introduce EvoGPT-f -- an evolutionary framework to quantify the differential machine learnability of these formal math languages. Here, we will treat the respective formal languages \emph{as the environment} in which an evolutionary process must take place, where natural selection forces populations of Transformers \cite{AttentionIsAllYouNeed} to optimize their training hyperparameters and architectures to best learn the statistical structure of the language. By systematically evaluating and normalizing the predictive power across languages and tokenization methods, we can begin to explore a proof-of-concept to benchmark the machine learnability of the respective formal math languages and forge new quantitative and qualitative comparative research that can act as a rising tide across all formalization communities.

Contributions of this paper:
\begin{itemize}
\item  1st systematic study aimed at benchmarking the machine learnability of formal math languages -- 5 formal math corpora (Lean 3, Lean 4, Coq, HOL 4, HOL Light) using 4 tokenization methods (character, word-level, Byte Pair Encoding \cite{bpe1, bpe2} and StarCoder \cite{li2023starcoder} tokenizer).
\item  A general system for ingesting and preparing corpora, running evolutionary pre-training experiments, and comparative analysis while creating formalization copilots. The data and model management (LLMOps) is custom designed, and can be toggled to local or cloud storage (AWS). A Streamlit application, EvoEDA, is provided for dynamic exploration of training results within and across evolutionary simulations, as well as high-level summary statistics (corpora, tokenization rounds, tokens, training rounds, etc) and inference experiments.
\end{itemize}

The remainder of this paper is organized as follows. Section II provides a brief introduction to formal mathematics (unfamiliar readers can find additional introductory material in Appendix ~\ref{appendix:a}), but quickly transitions into the respective corpora and data preparation. Section III introduces the EvoGPT-f algorithm, and Section IV on the validation and benchmarking results. In the discussion, we turn to the many areas of improvement and extension of these initial results.  

\section{Formal Mathematics Corpora}
\label{sec:headings}

This study is focused on the three most widely used formal mathematics languages -- Lean, HOL, and Coq --  and their various incarnations, for a total of five corpora (Lean 3, Lean 4, Coq, HOL 4, HOL Light) using four tokenization methods (character level, word level, Byte Pair Encoding (BPE), and StarCoder tokenizer). These languages are founded on type theory, enabling rigorous and precise representation of mathematical objects and theorems. Lean and Coq are based on the Calculus of Constructions  \cite{CalculusOfConstructions1988}, whereas HOL is founded on Church's theory of types \cite{ChurchSimpleTheoryTypes1940}. For those unfamiliar with formal math, an example formal theorem and proof -- the irrationality of the square root of 2 -- is provided in Appendix~\ref{appendix:a} for illustration and comparison among languages. 

Each corpus was processed similarly, with custom processing steps based on the nuances of the language. For example, Lean block comments are defined by $\text{/- ... -/}$, whereas Coq and HOL Light use $\text{(* ... *)}$. Table.~\ref{tab:table1} provides statistics on data preparation. Training to validation split is 80\% /\ 20\%. 

\begin{table}[ht]
\centering
\begin{tabular}{|c||c|c|c|c|c|}
\hline
Corpus & Files Processed & File Type & Corpus Size (chars) & Download Date & GitHub Link \\
\hline
\rowcolor{lightgrey}
Lean 3 & 3,401 & .lean & $\sim$35M & 2023-12-26 & \href{https://github.com/leanprover-community/mathlib}{Mathlib3 GitHub} \\
Lean 4 & 3,890 & .lean & $\sim$48M & 2023-12-26 & \href{https://github.com/leanprover-community/mathlib4}{Mathlib4 GitHub} \\
\rowcolor{lightgrey}
Coq  & 560 & .v & $\sim$5.5M  & 2023-12-26 & \href{https://github.com/coq/coq}{Coq GitHub} \\
HOL Light & 515 & .ml & $\sim$35M & 2023-12-26 & \href{https://github.com/jrh13/hol-light}{HOL Light GitHub} \\
\rowcolor{lightgrey}
HOL 4 & 2,486 & .sml & $\sim$70.5M & 2023-12-26 & \href{https://github.com/HOL-Theorem-Prover/HOL}{HOL4 GitHub} \\
\hline
\end{tabular}
\vspace{1mm}
\caption{Data Processing Statistics\label{tab:table1}}
\end{table}
Before running systematic evolutionary simulations using EvoGPT-f, a preliminary study is required to select the respective vocabulary sizes for BPE tokenization of each formal language corpus. Similar to the ``elbow method" used in k-means clustering, I used a fixed architecture to select an optimal vocabulary size for each corpus. For each corpus, ten equally spaced (on a log scale) vocab sizes are used for BPE tokenization. The bounds of the respective corpus BPE search space are the vocabulary sizes of character, and word-level tokenization, respectively. 

For each corpus and BPE vocab size, a GPT is trained for $n=1,000$ learning iterations and the average validation loss is estimated from $n=30$ bootstraps. The best validation loss from these evaluations is adjusted by the baseline entropy of the tokenization, where the baseline is the Shannon entropy $H = -\sum_{i=1}^{n} p(i) \log_2 p(i)$, where $p(i)$ is the relative frequency of token $i$.

The BPE search space for each corpus, and the corresponding adjusted validation loss, is shown in Appendix ~\ref{appendix:b}. Generally, the adjusted losses stabilize consistently when reaching a vocabulary size between $2,000$ to $4,000$, and this is how the final BPE vocabulary size is selected for each corpus. These final BPE vocabulary sizes, along with the other tokenization methods, are provided for all corpora in Table ~\ref{tab:table2}.

\begin{table}[ht]
\centering
\begin{tabular}{|c|c|c|c|}
\hline
\textbf{Corpus Type} & \textbf{Tokenization Method} & \textbf{Vocabulary Size} &  \textbf{Baseline Entropy} \\
\hline
\rowcolor{lightgrey}
        Coq &           Character &             131 &                 3.456047 \\
        Coq &                 BPE &           3,904 &                 5.869857 \\
\rowcolor{lightgrey}
        Coq &           StarCoder &           5,475 &                 5.689725 \\
        Coq &                Word &          21,320 &                 5.203272 \\
\hline
\rowcolor{lightgrey}
      HOL 4 &           Character &             257 &                 3.411286 \\
      HOL 4 &                 BPE &           4,555 &                 6.460252 \\
\rowcolor{lightgrey}
      HOL 4 &           StarCoder &          17,159 &                 6.211086 \\
      HOL 4 &                Word &         165,655 &                 5.447621 \\
\hline
\rowcolor{lightgrey}
  HOL Light &           Character &             148 &                 3.734230 \\
  HOL Light &                 BPE &           2,091 &                 6.069714 \\
\rowcolor{lightgrey}
  HOL Light &           StarCoder &          10,315 &                 5.682008 \\
  HOL Light &                Word &          57,324 &                 5.171080 \\
\hline
\rowcolor{lightgrey}
     Lean 3 &           Character &             421 &                 3.532371 \\
     Lean 3 &                 BPE &           2,673 &                 5.965266 \\
\rowcolor{lightgrey}
     Lean 3 &           StarCoder &          10,702 &                 5.481840 \\
     Lean 3 &                Word &         107,756 &                 5.602466 \\
\hline
\rowcolor{lightgrey}
     Lean 4 &           Character &             513 &                 3.625152 \\
     Lean 4 &                 BPE &           3,566 &                 6.263405 \\
\rowcolor{lightgrey}
     Lean 4 &           StarCoder &          13,524 &                 5.854967 \\
     Lean 4 &                Word &         172,410 &                 5.923884 \\
\bottomrule
\end{tabular}
\vspace{1mm}
\caption{Vocabulary Size and Baseline Entropy by Corpus and Tokenization Method\label{tab:table2}}
\end{table}

\section{EvoGPT-f Framework}

Evolutionary algorithms are a class of search and optimization algorithms that use mutation and cross-over genetic events found in biological evolution. The EvoGPT-f Framework is a genetic algorithm that evolves populations of Transformers, treating the formal language \emph{as the environment} in which the evolutionary process must take place. The ability of the genetic algorithm to evolve more fit GPT populations provides a quantitative framework for comparing the difficulty of learning these respective formal languages by neural language models. 

EvoGPT-f is written in Python using the PyTorch \cite{NEURIPS2019_9015} library, and the nanoGPT Transformer model definition from \href{https://github.com/karpathy/nanoGPT/blob/master/model.py}{nanoGPT GitHub} was used as a seed to the design with nominal changes (e.g. lowering the standard deviation of the normally distributed initial weights, to mitigate the risk of exploding gradients due to the relatively smaller corpora). It is feasible to run \emph{small}-scale local testing of these experiments, such as on a MacBook Pro M1 with a single GPU, 10 cores and 32Gb of memory. The results herein where computed using 8 NVIDIA Tesla V100 GPUs with 16GB of HBM2 VRAM each (5120 CUDA cores per card, 640 Tensor cores per card). I design a custom LLMOps system for EvoGPT-f, with Amazon Web Services (AWS) S3 and Relational Data Service (RDS) Postgres database on the backend for raw corpora data, tokenization, and training metadata management and governance.

There are a set of training round parameters, GPT parameters tuned by natural selection, and fixed parameters for all evolutionary simulations. This complete list of parameters, class and bounds (if applicable) are provided in Table ~\ref{table:model_params}.

\begin{table}[H]
\centering
\begin{tabular}{ | m{3cm} | m{2cm} | m{3cm} | m{7cm} | }
\hline
\textbf{Parameter} & \textbf{Type} & \textbf{Bounds} & \textbf{Description} \\
\hline
\rowcolor{lightgrey}
\emph{n\_generations} & Training Round & NA & Number of generations to evolve. \\
\hline
\emph{population\_size} & Training Round & NA & The number of Transformers per generation. \\
\hline
\rowcolor{lightgrey}
\emph{max\_iters} & Training Round & NA & Total number of training iterations. \\
\hline
\emph{elite\_percent} & Training Round & NA & The \% of top ranked Transformers to create the next generation. \\
\hline
\rowcolor{lightgrey}
\emph{crossover\_probability} & Training Round & NA & The probability of a crossover event. \\
\hline
\emph{mutation\_probability} & Training Round & NA & The probability of a mutation event. \\
\hline
\rowcolor{darkgrey} & & & \\ 
\hline
\rowcolor{lightgrey}
\emph{n\_layer} & Evo. Tuning & [2,12], dependent on n\_embd.  & The number of layers in the transformer model. \\
\hline
\emph{n\_heads} & Evo. Tuning & $2^{i}$ for $i=0,...,4$, but must be a factor on n\_embd. & The number of attention heads in each layer. \\
\hline
\rowcolor{lightgrey}
\emph{n\_embd} & Evo. Tuning & Typical training runs were $2^{i}$ for $i=5,...,9$ & The dimensionality of the embeddings for each input token. \\
\hline
\emph{batch\_size} & Evo. Tuning & Typical training runs were $2^{i}$ for $i=1,...,8$ & The number of training samples used in each training step. \\
\hline
\rowcolor{lightgrey}
\emph{block\_size} & Evo. Tuning & Typical training runs were $2^{i}$ for $i=3,...,11$ & The maximum number of tokens in a sequence. \\
\hline
\emph{dropout} & Evo. Tuning & [0.0, 0.5] & The probability of dropping out a neuron --randomly and temporarily-- in each layer during training. \\
\hline
\rowcolor{lightgrey}
\emph{learning\_rate}, \emph{min\_lr} & Evo. Tuning & [$1\times10^{-4}$,$1\times10^{-3}$] and [$1\times10^{-5}$,$1\times10^{-4}$] & The learning\_rate is the step size for updating the model parameters during each iteration of the optimization, and the min\_lr is its lower bound. \\
\hline
\emph{beta1}, \emph{beta2} & Evo. Tuning & [0.1, 0.98] and [0.1, 0.99] & Used in AdamW optimizer \cite{AdamWOptimizer}, these control the exponential decay rates for the first and second moments of the gradients. \\
\hline
\rowcolor{lightgrey}
\emph{weight\_decay} & Evo. Tuning & [0, 0.1] & The decoupled weight decay (also introduced in \cite{AdamWOptimizer}) used for regularization. \\
\hline
\rowcolor{darkgrey} & & & \\
\hline
\rowcolor{lightgrey}
\emph{eval\_iters} & Fixed & NA & Bootstrap training and validation loss estimate every eval\_iters. Set to $n=10$ for experiments. \\
\hline
\emph{decay\_lr} & Fixed & NA & Boolean indicating whether to decay the learning rate. Set to True for experiments. \\
\hline
\rowcolor{lightgrey}
\emph{warmup\_iters} & Fixed & NA & Number of iterations to linearly ramp up the learning rate at the start. Set to 50\% of max\_iters for experiments. \\
\hline
\emph{lr\_decay\_iters} & Fixed & NA & Number of iterations over which to decay the learning rate. $\sim $max\_iters for all experiments. \\
\hline
\end{tabular}
\vspace{1mm}
\caption{Framework and Model Parameters \label{table:model_params}}
\end{table}

\begin{algorithm}[ht]
\label{algo:evogptf_algo}
\caption{EvoGPT-f} 
\begin{algorithmic}[1]
        \For {(formal\_language, tokenization\_method) in Table ~\ref{tab:table2}}
            \State Load training and validation data for formal\_language based on tokenization\_method
    
            \State initial\_pop = []
            \For {$i=1, \ldots$, population\_size}
                \State rand\_chrom = get\_random\_chromosome()
                \State initial\_pop.append(rand\_chrom)
            \EndFor
    
            \For {$g=1,\ldots,$ n\_generations}
                \State if g == 1:
                    \INDSTATE{2em}  current\_pop = initial\_pop
                \State else:
                    \INDSTATE{2em}   current\_pop = next\_gen
                \For {c in current\_pop}
                    \State fitness = train\_gpt(c)
                \EndFor
    
                \State Rank best\_val\_loss
                \State elite\_pop = elite\_perc of GPT's 
                
                \State next\_gen = []
                \While {len(next\_gen) $<$ population\_size}
                    \State parent1, parent2 = sample(elite\_pop, 2, replace=F)
                    \State offspring = get\_two\_offspring(parent1, parent2, crossover\_probability)
                    \For {o in offspring}
                        \State if rand\_prob $<$ mutation\_probability:
                            \INDSTATE{2em} o $\leftarrow$ mutate\_chromosome(o)
                    \EndFor
                    \State  next\_gen.append(offspring)
                \EndWhile
                
            \EndFor
            \State Save simulation round, population level, and training iteration level metadata and performance
        \EndFor
\end{algorithmic} 
\end{algorithm}

A member of the population constitutes a Transformer with specific architecture and training hyperparameters. The EvoGPT-f mutation and crossover mechanism are customized for the transformer hyperparameters and architecture. For example, a mutation event in the learning rate is as follows. Let the learning rate mutation step size be defined as $\delta_{\text{lr}} = \frac{1\times10^{-3} - 1\times10^{-4}}{20}$ Then, a random +/- mutation from the current learning rate $\text{lr}_{0}$ is given by:
\[ \text{lr}_{1} = \text{clip}\left(\text{lr}_{0} \pm \delta_{\text{lr}}, 1\times10^{-4}, 1\times10^{-3}\right) \]
where \(\text{clip}(x, a, b)\) clips \(x\) within the range \([a, b]\).

Given the memory constraints in the aforementioned GPU cluster, the space of computationally feasible architectures -- i.e. (n\_layer, n\_head, n\_embd) triples -- are pre-computed in a data frame. When the architecture is randomly selected to be mutated, the mutation mechanism finds the adjacent rows of the current architecture and selects, with equally probability, from the two closest neighbors; thereby randomly walking the data frame rows during mutation.

\section{EvoGPT-f Results}

There are two phases of results: model and framework validation, and differential machine learnability results. These results are based on an evolutionary simulation conducted for $n\_generations = 15$, with $population\_size = 10$, for a total of $n=150$ GPTs trained and evaluated. This was one of $\sim30$ evolutionary simulations which informed the observations. 

I will reiterate here that these results -- specifically the differential learnability benchmarking in section ~\ref{sec:results2} -- are not intended as definitive claims and ranks of the respective formal languages with regards to language design or the resulting ease at which machines or humans can learn them.  

\subsection{Model and Framework Validation}
\label{sec:results1}
Here, we validate the individual and collective GPT performance as well as the expected evolutionary behavior.

\begin{figure}[H]
    \centering 
    \includegraphics[width=0.75\textwidth]{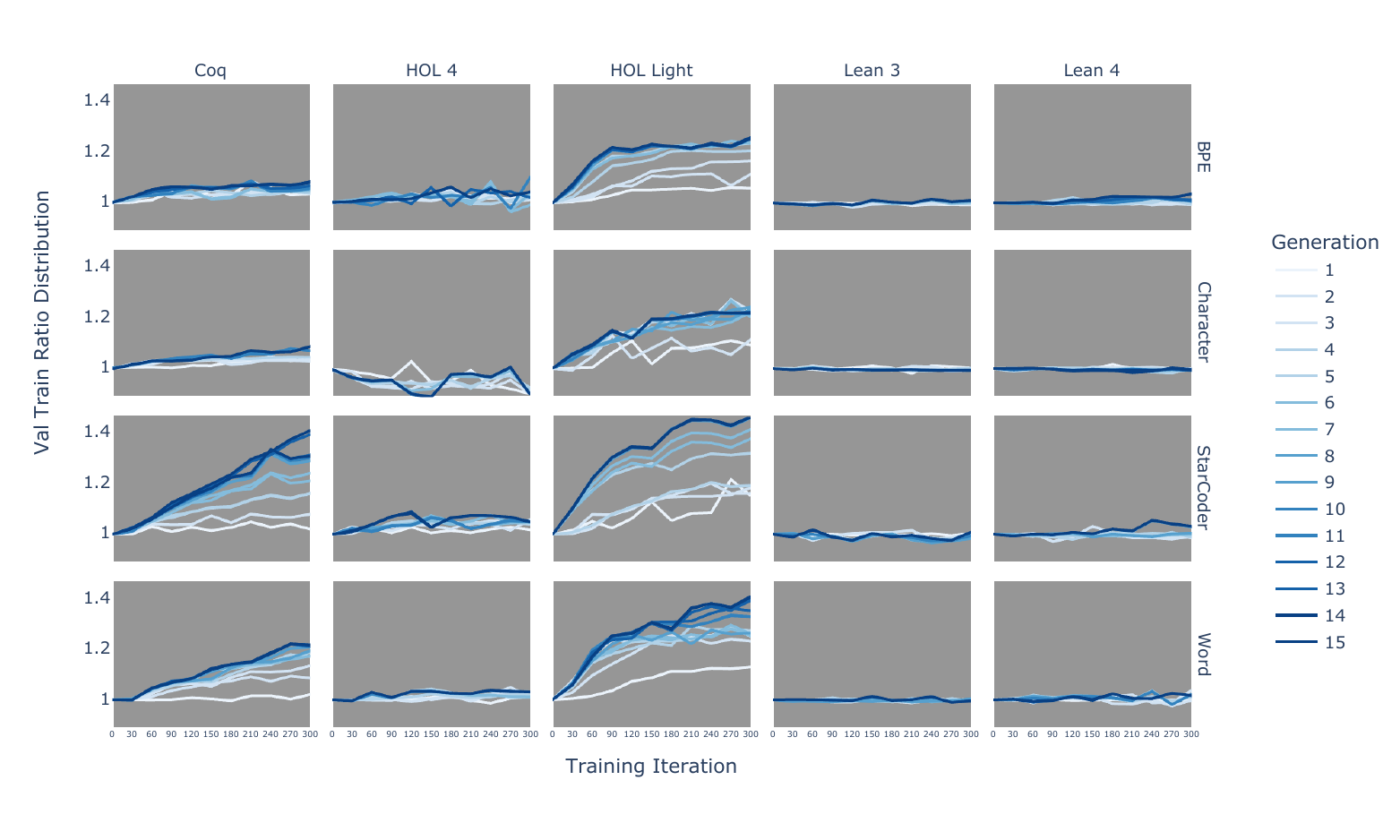}
    \caption{ Validation to Training Loss Ratio across Generations}
    \label{fig:val_train_ratio_line_plot_by_train_iter_cf_corpus_rf_tkn}
     
    \includegraphics[width=0.75\textwidth]{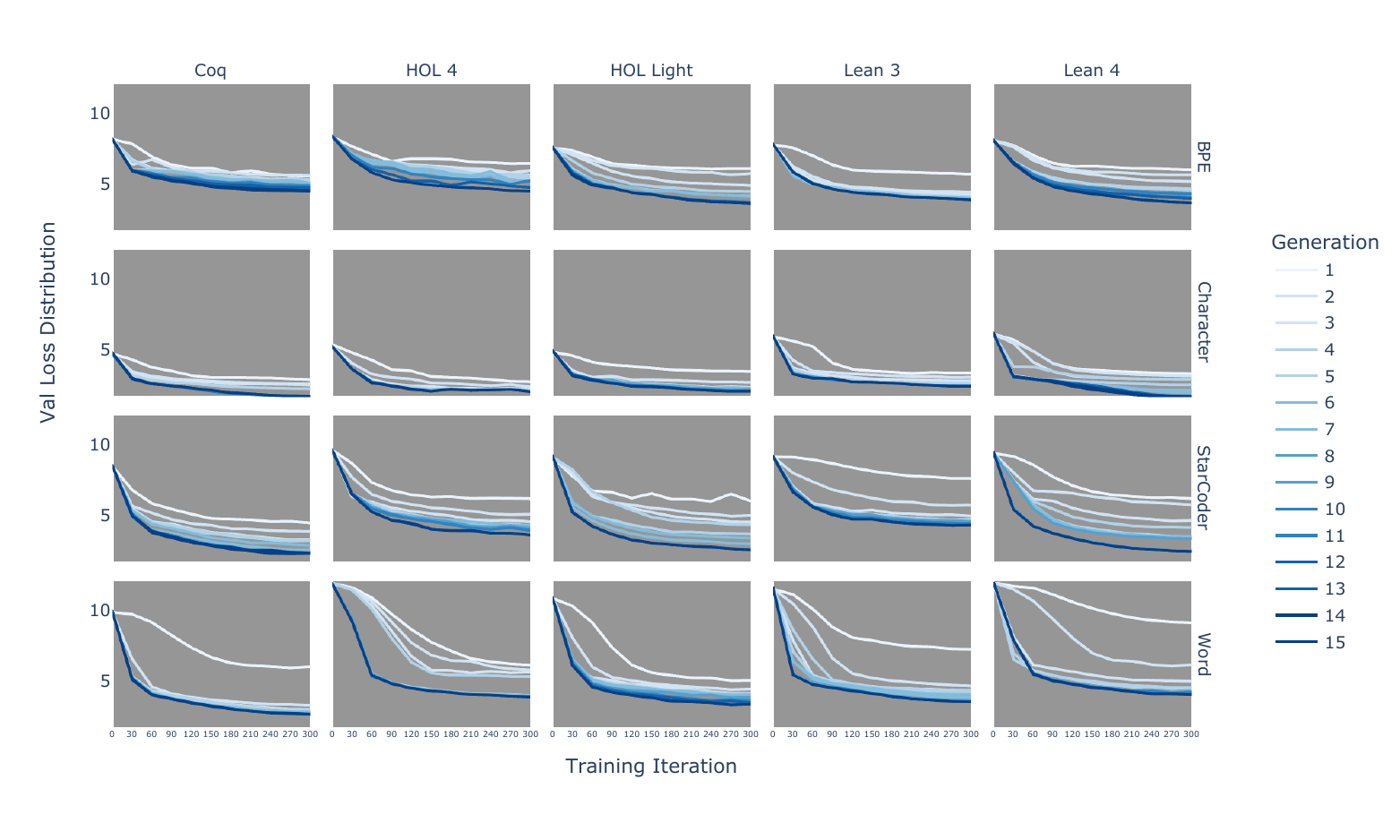}
    \caption{Validation Loss across Generations}
    \label{fig:val_loss_line_plot_by_train_iter_cf_corpus_rf_tkn}
\end{figure}

These corpora are relatively smaller than conventional human language modeling datasets, and thus present greater risk of overfitting when searching larger architectures. Figure ~\ref{fig:val_train_ratio_line_plot_by_train_iter_cf_corpus_rf_tkn} shows the $valid/train$ loss ratio by corpus and tokenization method across training iterations (using the mean validation and training loss from $n=30$ bootstrap samples, measured every 10 training iterations). We observe: 1. parity across tokenization methods for both Lean 3 and Lean 4; 2. mild to strong parity of HOL 4, albeit with elevated volatility for some tokenization methods; 3. some generational drift from parity for Coq using Word and StarCoder tokenization; 4. significant evolutionary departure of validation from training (20-40\%) for HOL Light systematically across tokenization methods. The latter was to be expected, based on the relatively smaller size of the HOL Light corpus. The departure of $valid/train$ from unity will be used as part of the weighting method for normalizing corpus learnability across tokenization methods; as will the baseline entropy of the vocabulary of the tokenization method. 

Turning to the validation of the evolutionary process, we observe in Figure  ~\ref{fig:val_loss_line_plot_by_train_iter_cf_corpus_rf_tkn} the strict dominance (in game-theoretic terms) of the validation loss curves of latter generations over earlier generations; this behavior is consistent across corpora and tokenization methods.

We also want to inspect the correlation between hyperparameters and: 1. validation loss, to ensure it agrees with intuition and to gain insights into training on these corpora and tokenization methods; 2. generation, for additional evidence that the evolutionary tuning is behaving as expected. 

\begin{figure}[H]
    \centering 
    \includegraphics[width=1.05\textwidth]{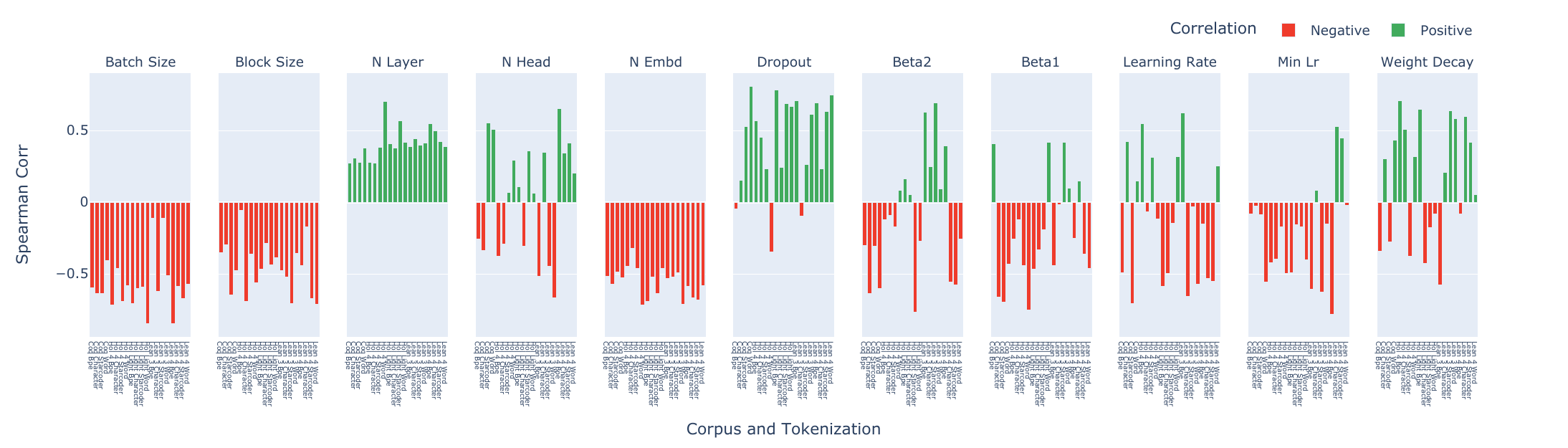}
    \caption{Spearman Correlation between GPT Hyperparameters and Validation Loss}
    \label{fig:corr_bar_chart_hyperparm_vs_val_loss}

    \vspace{2mm}

    \includegraphics[width=1.05\textwidth]{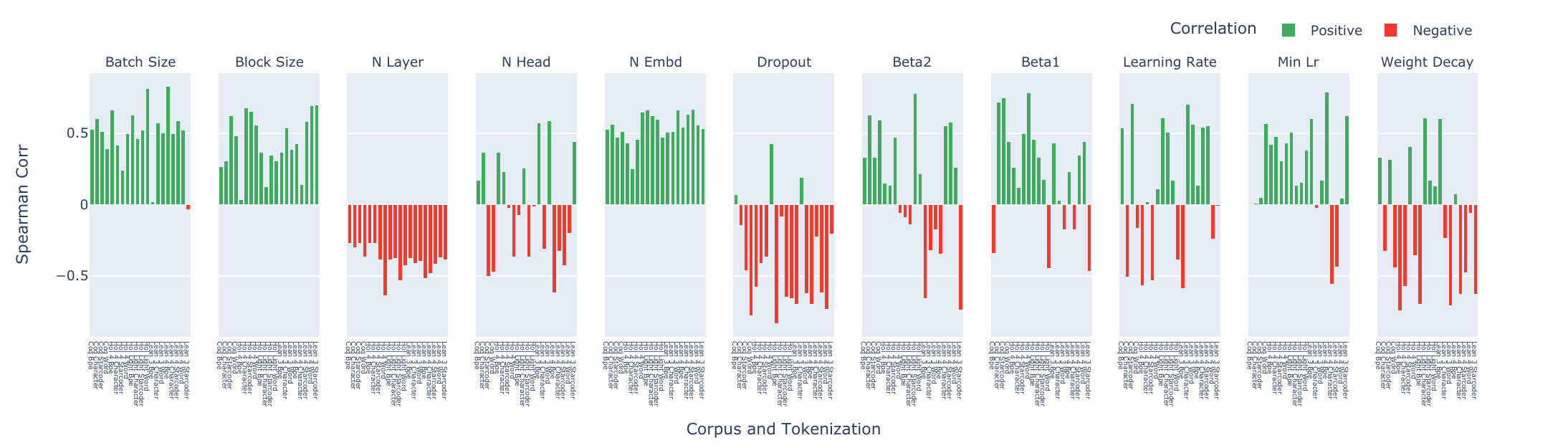}
    \caption{Spearman Correlation between GPT Hyperparameters and Generation}
    \label{fig:corr_bar_chart_hyperparm_vs_Generation}
\end{figure}

Figure ~\ref{fig:corr_bar_chart_hyperparm_vs_val_loss} shows generally intuitive correlation between validation loss and hyperparameters, such as improved loss for larger batch/block size, and embedding dimension. However, although the number of embeddings consistently improve model performance, deeper networks (n\_layer) systematically do not generalize well, and the benefit of greater parallelization of attention is inconsistent. We observe these nuanced and relatively smaller corpora are sensitive to regularization, evidenced by the systematic positive correlation with dropout percentage. We also observe nearly a systematic negative correlation with the AdamW beta1, suggesting a preference with weighting past gradients (i.e., smoothing), to mitigate overfitting while learning these relatively smaller and specialized corpora.

\begin{figure}[ht!]
    \centering 
    \includegraphics[width=0.8\textwidth]{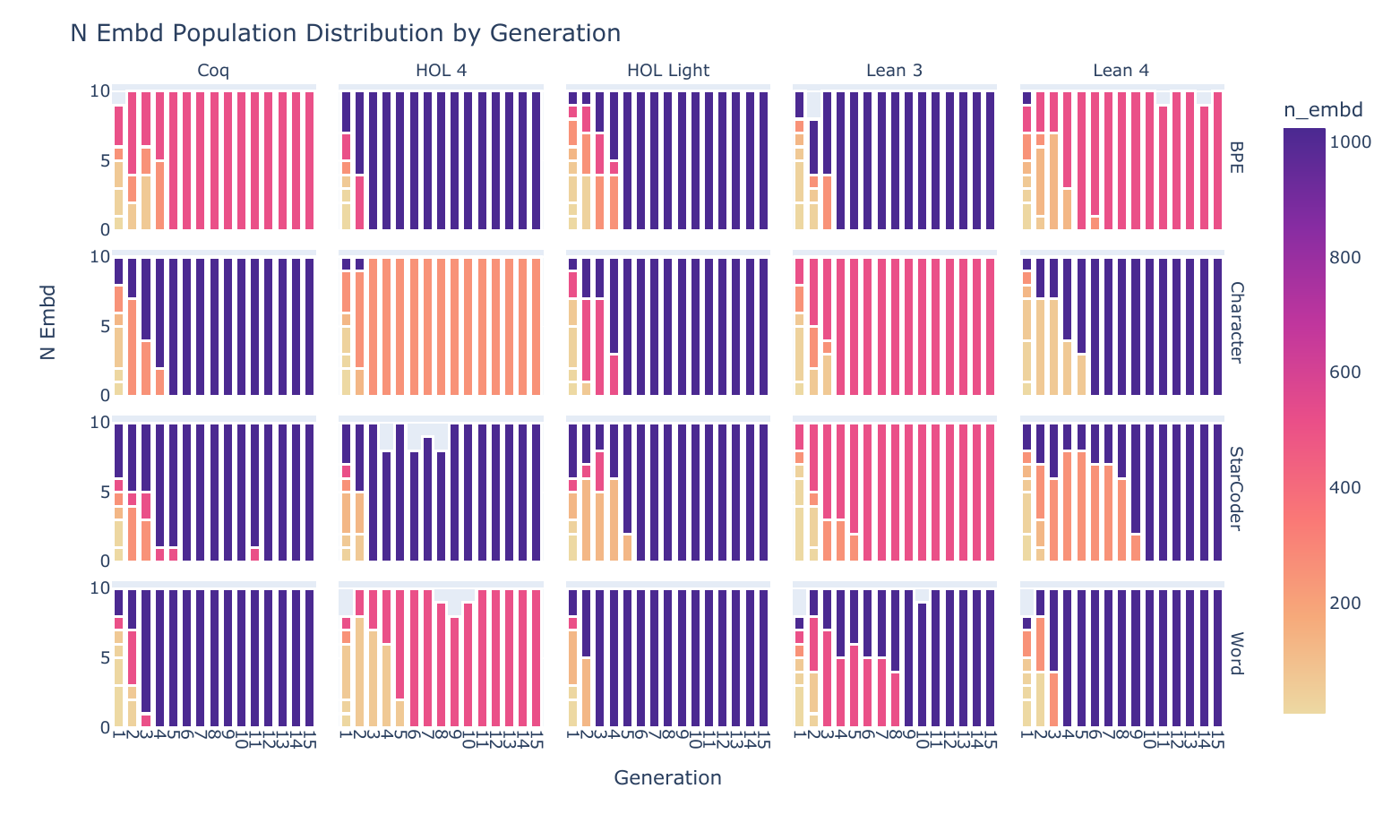}
    \caption{Selection for Larger Embedding Dimension}
    \label{fig:corr_gen_n_embd_moved}
    \vspace{1mm}

    \includegraphics[width=0.8\textwidth]{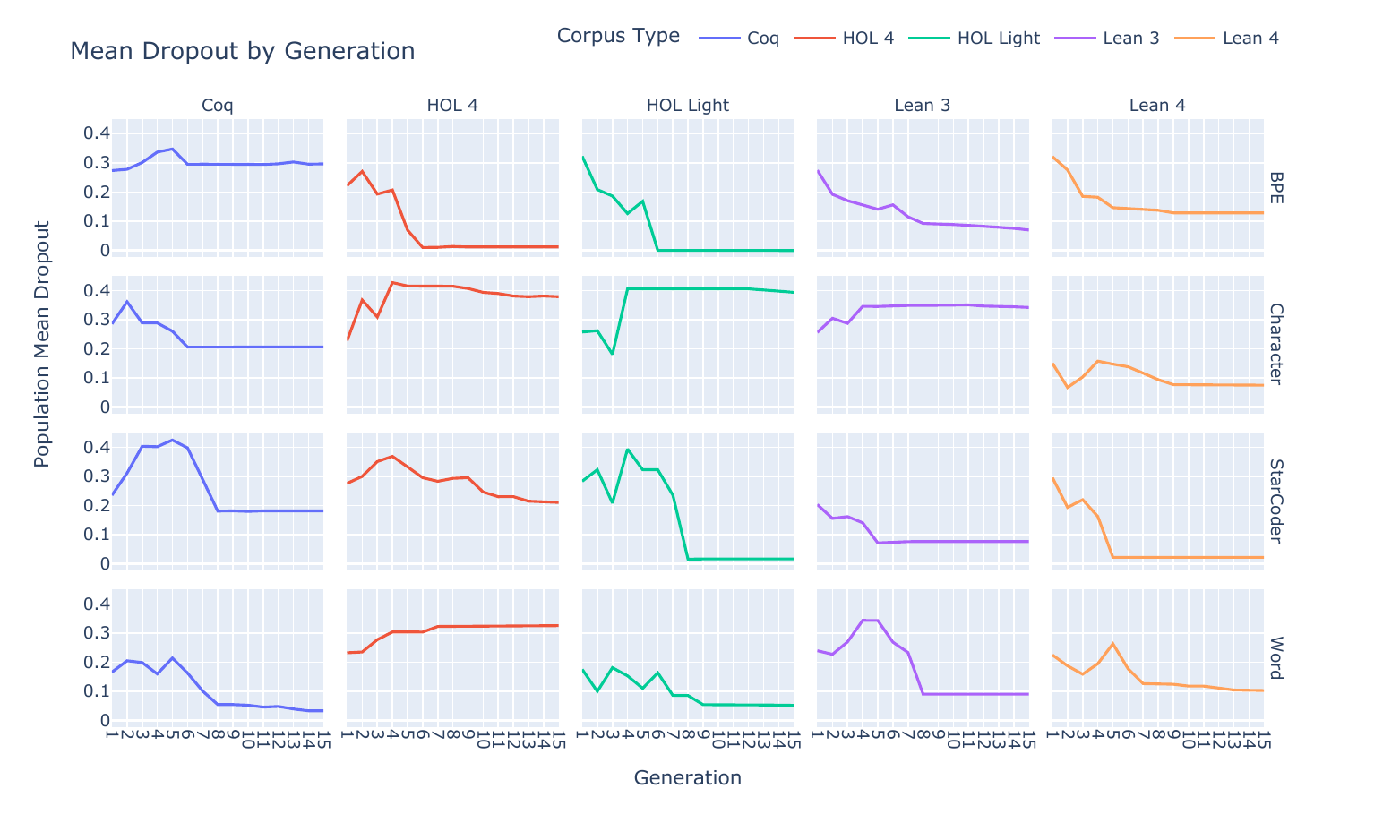}
    \caption{Selection for Lower Dropout \%}
    \label{fig:corr_gen_dropout_moved}
    
\end{figure}

Figure ~\ref{fig:corr_bar_chart_hyperparm_vs_Generation} shows the correlation of these hyperparameter values with the generation of the evolutionary process. These generally mirror the correlations with validation loss we see above, such as selection pressure for larger batch/block, embedding dimension, learning rates and beta1 values. Similarly, we observe a negative selection for n\_layer and dropout percentage. Examples of the underlying bivariate correlations are provided in Figures ~\ref{fig:corr_gen_n_embd_moved} and ~\ref{fig:corr_gen_dropout_moved}, and dynamic exploration in EvoEDA is illustrated in Appendix ~\ref{appendix:c5}.

\subsection{Benchmarking Formal Language Learnability}
\label{sec:results2}
To synthesize GPT performance across tokenization methods and compare at the corpora level, I provide two complementary methods for visual and statistical analysis. Both methods incorporate two aspects: penalizing the discrepancy between training and validation losses, and adjusting for the baseline entropy of the corpus. 

\begin{figure}[ht!]
    \centering 
    \includegraphics[width=0.8\textwidth]{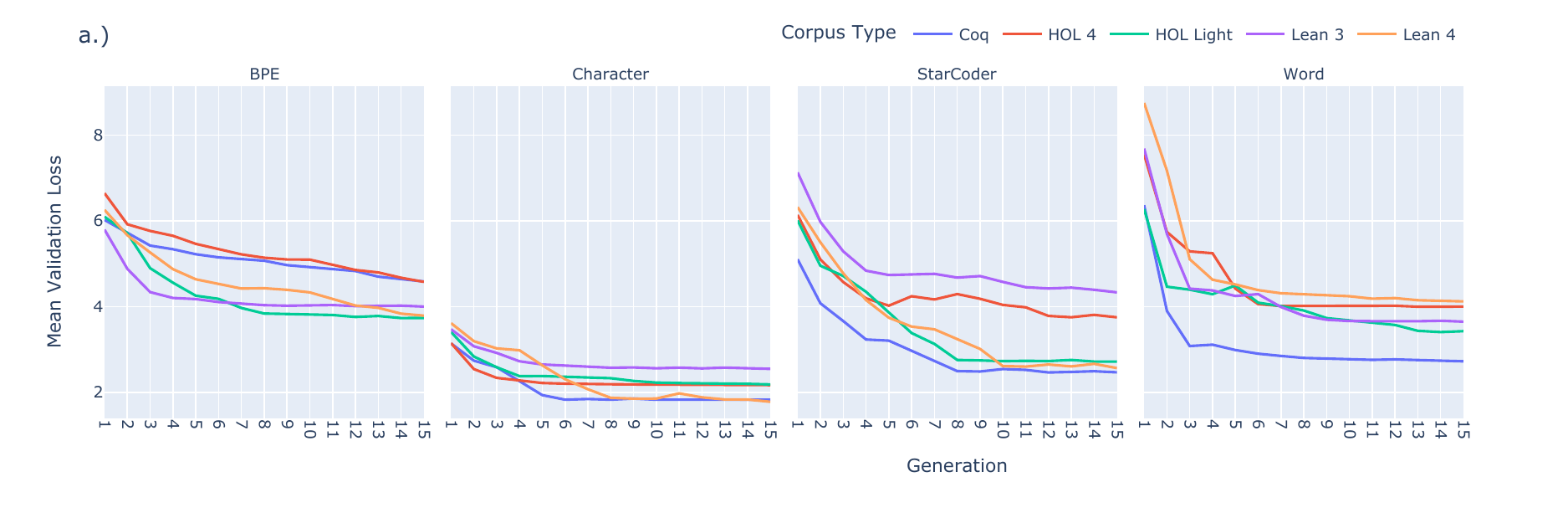}
    \label{fig:line_plot0_best_val_loss_by_generation_cf_tkn}
    \vspace{1mm}

    \includegraphics[width=0.8\textwidth]{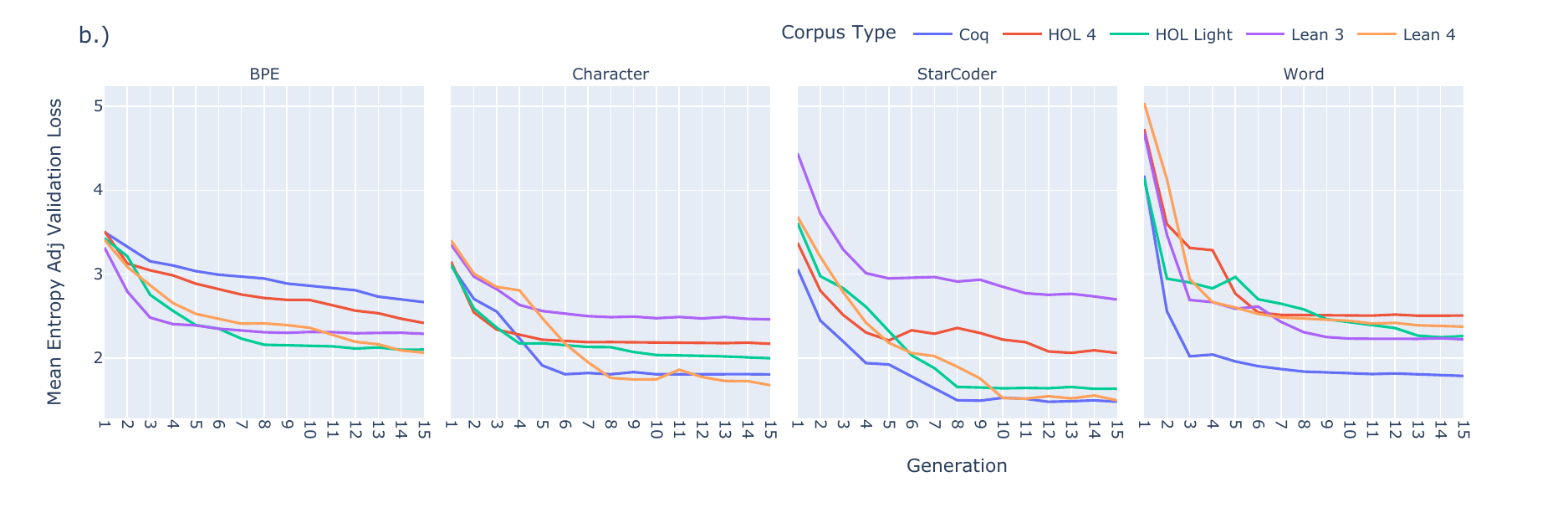}
    \label{fig:line_plot1_entropy_adj_val_loss_by_generation_cf_tkn}
    \vspace{1mm}

    \includegraphics[width=0.8\textwidth]{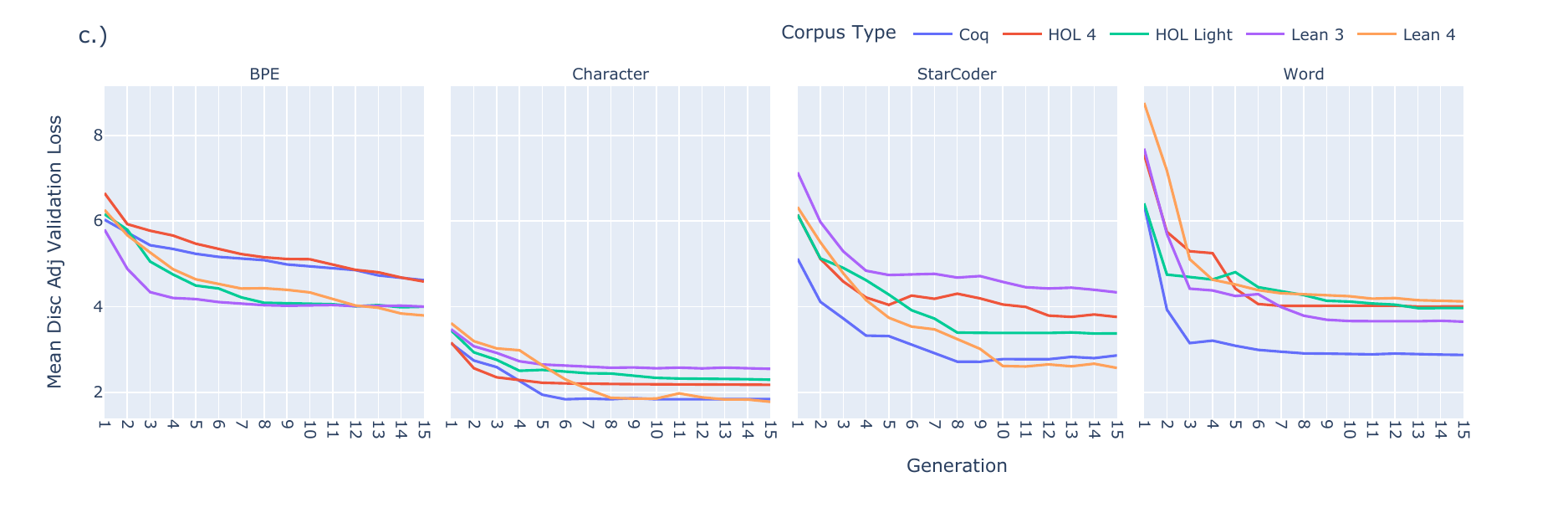}
    \label{fig:line_plot2_discrepancy_adj_val_loss_by_generation_cf_tkn}
    \vspace{1mm}

    \includegraphics[width=0.8\textwidth]{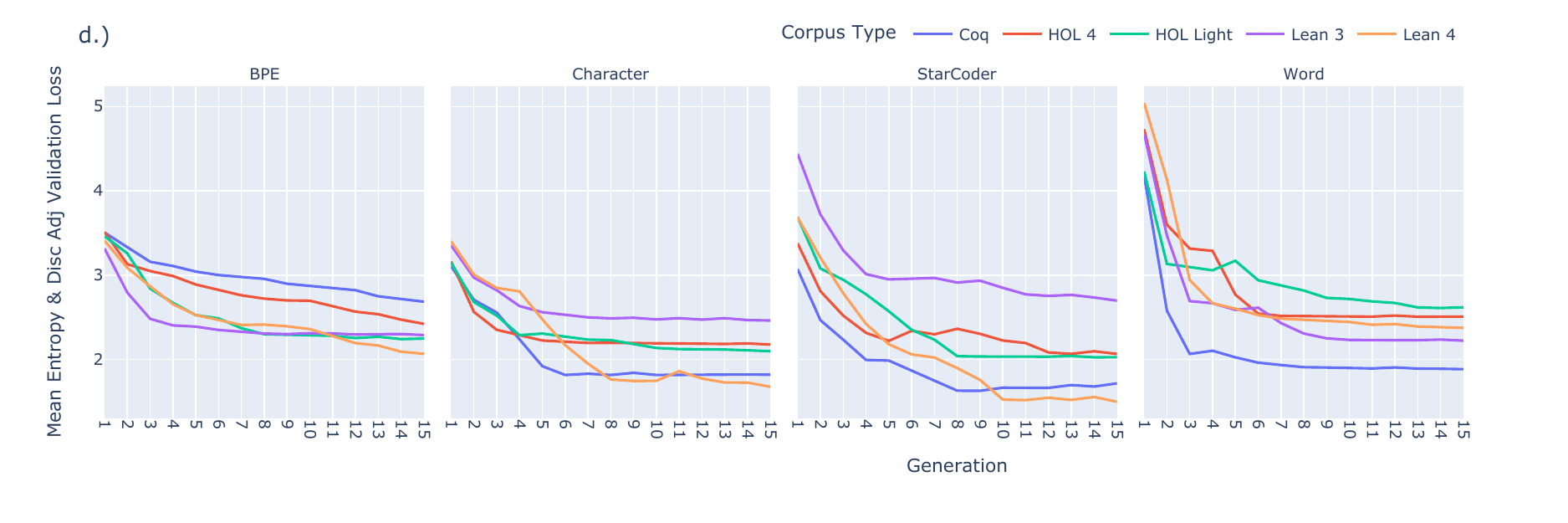}
    \caption{a.) Original Validation Loss, b.) Entropy Adjusted Validation Loss, c.) Overfitting Penalty Adjusted Validation Loss, d.) Final Adjusted Validation Loss}
    \label{fig:line_plot3_entropy_and_discrepancy_adj_val_loss_by_generation_cf_tkn}
    
\end{figure}

\textbf{Method 1: Calibration for Visualization.} Here, we will calibrate the original validation loss with two factors. Let $L_{c,k}^v$ and $L_{c,k}^t$ be the validation and training loss, respectively, for a given corpus $c$ and tokenization method $k$. Let $H_{c,k}$ be the Shannon entropy of the tokens for a given $(c,k)$. The normalized validation loss $L_{c,k}^{n_{1}}$ is defined by:
\begin{equation}
    L_{c,k}^{n_{1}} = L_{c,k}^v * \left( \frac{1}{ \exp\left(-(\frac{L_{c,k}^v}{L_{c,k}^t}-1)^{2}\right)} \right) * \left(\frac{\min\limits_{c,k}H_{c,k}}{H_{c,k}} \right) 
\end{equation}
The first Gaussian component is an overfitting penalty which amplifies the loss based on the discrepancy between the model's training and validation loss. The second component adjusts for the relative entropy level (i.e. inherent prediction difficulty) of the corpora and tokenization method using the entropies in Table ~\ref{tab:table2}. This transformation process, from original validation loss, to the application of each adjustment, and then the final normalized validation loss are shown in Figure ~\ref{fig:line_plot3_entropy_and_discrepancy_adj_val_loss_by_generation_cf_tkn}. The underlying distributions are provided in Appendix ~\ref{appendix:c1} (grouped by tokenization method) and Appendix ~\ref{appendix:c2} (grouped by corpus). By comparing Figure 5a to Figure 5d, we see that the normalization both: 1. shifts corpus curves to be better aligned across respective tokenization methods; 2. provides greater segmentation of real performance, evidenced by the relative improvements in Lean 4 and the regression of HOL Light performance.

\textbf{Method 2: Difference from Baseline Entropy}
The first `Calibrated Loss' given by $L_{c,k}^{n_{1}}$ enabled us to visualize the evolution at the corpora level. We want to complement this approach with a variety of statistical models to test for a difference in expected normalized loss across corpora. So I introduce an alternative normalized loss, given by $L_{c,k}^{n_{2}} = L_{c,k}^v - H_{c,k}$, referred to as `Information Gain' in modeling results. 

The response $L_{c,k}^{n_{2}}$ has a Gamma structure, aside from the support being not strictly positive. Therefore, we can model using standard Ordinary Least Squares (OLS) with a tenuous normality assumption, or we can shift using $LS_{c,k}^{n_{2}} = L_{c,k}^{n_{2}} - min(L_{c,k}^{n_{2}}) + 0.01$, where the $0.01$ term is merely a small number added so that the $min(L_{c,k}^n)$ model observation is now strictly positive. The latter approach is chosen\footnote{This transformation, however, is at the expense of interpretation of the corpora factors, which is why we wanted the raw difference in the first place -- i.e. to express the results in terms of reduction in uncertainty from the baseline entropy.} so that a Gamma distribution can be assumed for $LS_{c,k}^{n_{2}}$. This transformation can be found in Appendix ~\ref{appendix:c4}. $LS_{c,k}^{n_{2}}$ is referred to as the `Shifted Information Gain' in the model results.

\begin{figure}[H]
    \centering
    \includegraphics[width=0.8\textwidth]{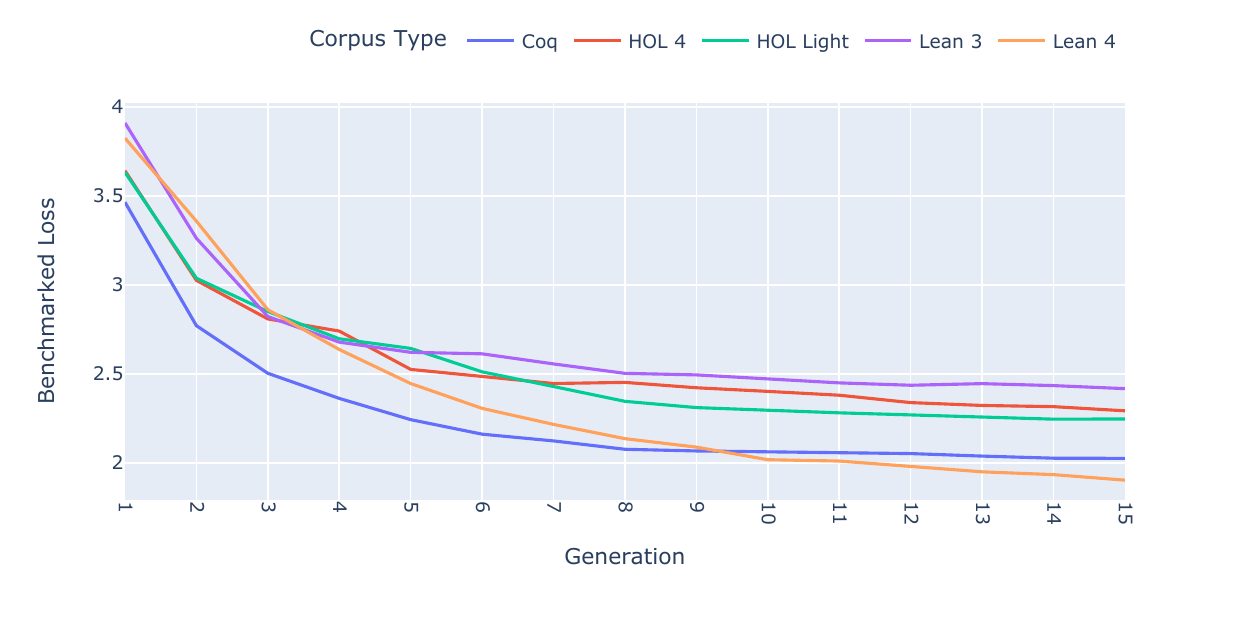}
    \caption{ Mean Calibrated Loss $L_{c}^{n_{1}}$ }
    \label{fig:entropy_and_discrepancy_adj_val_loss_final_by_Generation}
    \vspace{1mm}
    \includegraphics[width=0.8\textwidth]{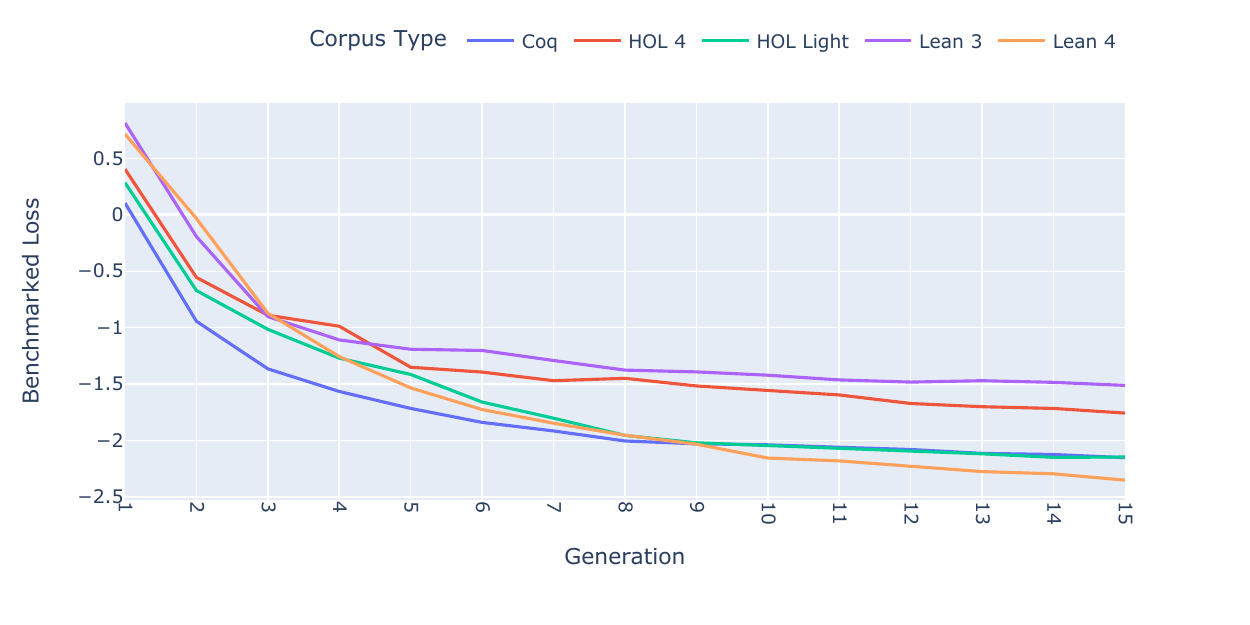}
    \caption{ Mean Information Gain $L_{c}^{n_{2}}$}
\label{fig:line_plot3_entropy_and_discrepancy_adj_val_loss_by_generation_cf_tkn_main}
\end{figure}

Figures ~\ref{fig:entropy_and_discrepancy_adj_val_loss_final_by_Generation} and ~\ref{fig:line_plot3_entropy_and_discrepancy_adj_val_loss_by_generation_cf_tkn_main} show the evolutionary trajectories of the mean normalized validation loss $L_{c}^{n_{1}}$ and mean Information Gain $L_{c}^{n_{2}}$, respectively, when aggregated to the corpus level across generations. Since the major departure of these two methods is the absence of the overfitting adjustment in $L_{c,k}^{n_{2}}$, we would expect generally similar behavior except for the HOL Light trajectory. This is indeed the case, as $L_{\text{HOL Light}}^{n_{1}}$ is relatively worse ranking than $L_{\text{HOL Light}}^{n_{2}}$.

\textbf{Regression Modeling}
The responses, $L_{c,k}^{n_{1}}$ and $LS_{c,k}^{n_{2}}$, are each used for three separate regression models. The first two models are Generalized Estimating Equations (GEE) with Gamma distribution and log-link, to test for a difference in normalized loss across corpora while accounting for the within corpus autocorrelation across generations. There are two variants because one model is weighted regression, where the aforementioned Gaussian weights (not inverted) are used to assign less credibility to those model observations when testing for differential learnability across corpora. The third model is a Generalized Linear Model (GLM) with Gamma distribution and log-link, to test for a difference in learnability across corpora in the final generation.

\begin{figure}[H]
    \centering 
    \includegraphics[width=1\textwidth]{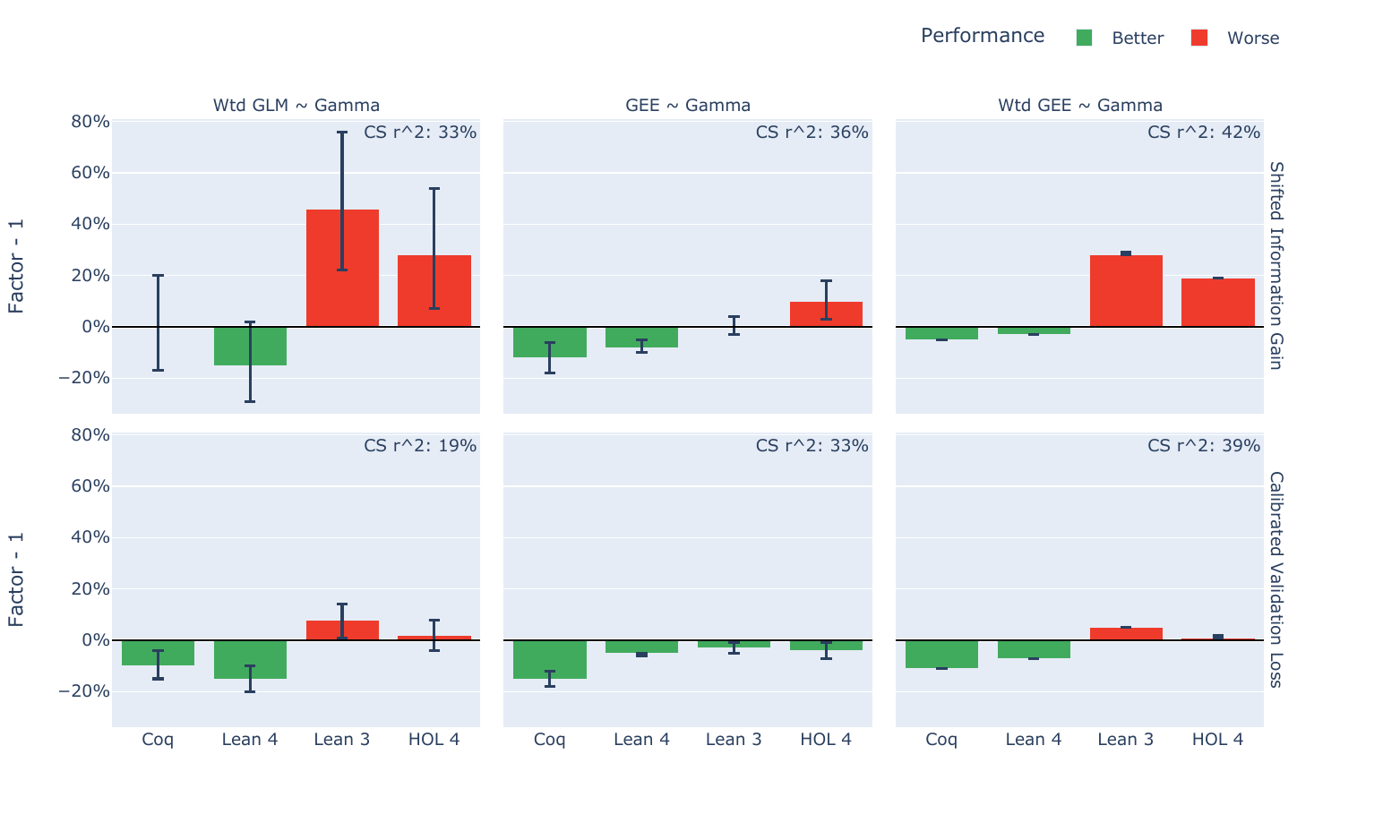}
    \caption{GEE and GLM Factors for Calibrated Loss and Shifted Information Gain Models}
    \label{fig:all_model_factors_plot}
\end{figure}

In Figure ~\ref{fig:all_model_factors_plot}, corpus factors are interpreted as the $100*(e^{\beta_{c}}-1)$ percent change in the response from the baseline corpus (HOL Light). The Cox-Snell likelihood ratio pseudo R-squared is provided in the upper-right for each model. Here, we see that the weighted models have greater predictive power in general; as do the GEE models, compared to the GLM used for testing performance differences in the final generation. The details of model parameter inference are provided in Appendix ~\ref{appendix:c4}. 

As suggested by Figures ~\ref{fig:entropy_and_discrepancy_adj_val_loss_final_by_Generation} and ~\ref{fig:line_plot3_entropy_and_discrepancy_adj_val_loss_by_generation_cf_tkn_main}, we find a significantly greater facility of adapted pre-training on Lean 4 and Coq, both from an evolutionary trajectory perspective and ranking in the final generation. Notably for the Lean community after their herculean effort to redesign and migrate from Lean 3 to Lean 4, we observe a significant improvement of $\sim20-50\%$ in machine learnability between these versions.


\section{Discussion}

It is a truly remarkable time to be working at the intersection of formal mathematics and artificial intelligence. The decades of blood, sweat and tears of formalization communities have culminated to, what many interpret as, an inflection point in formal mathematics in its quest to become mainstream. This is in part due to the seminal year of AI after the release of ChatGPT and its pervasive impact in re-envisioning the scientific research and discovery process. It is also in part due to the organizational maturity and growth of the Lean community and its awareness by the general public. These forces have converged and there is now a virtuous cycle of humans and machines working together to realize the full potential of formalization on the speed and verification of mathematical discovery. 

Now, more than ever, is the time for industry partners to expand their commitments to providing resources and capabilities that help forge cross-community research and development that may have been frustrated in the past due to ideological differences and institutional incentives.

I showed a poof-of-concept framework and initial results towards benchmarking the machine learnability of formal languages with an interdisciplinary approach from neural language models and evolutionary methods. This is not intended as a definitive claim with regards to the ``best" or ``easiest" language to learn. The statistical properties learned during pre-training autoregressive language models do not take into account qualitative or aesthetic properties of the language design, or the formalism of the mathematics itself. My hope is that these capabilities both motivate and provide the playground for more systematic quantitative and qualitative research across formal languages and communities.

To this end, there are many directions to continue improving and building on this work, and I welcome these enhancements when the Github repository is made public. In the meantime, please email if you have interest in these efforts.

\textbf{Additional tailored pre-processing}. Custom corpora pre-processing tailored to the nuances of the respective languages would be beneficial, including custom parsing and special tokens for respective languages. Furthermore, testing down/up-sampling of corpora to better align corpus sizes.

\textbf{Tokenization}. Tokenization is of paramount importance. Spanning tokenization methodologies and then model averaging was a means to mitigate any tokenization effect, as well as providing additional flexibility in the framework for future lines of inquiry. However, going beyond tokenization to the byte-level, such as in \cite{yang2023leandojo}, can also be added and explored. Furthermore, the preliminary BPE vocabulary size selection study was rudimentary, as the optimal vocabulary size is entangled with the architecture optimization. This can certainly be enhanced.

\textbf{Ontology and mapping}. Using the data engineering and analysis capabilities to further efforts of ontology building and mappings between formal languages, so that the search and comparison of mathematical objects and proofs can be more easily explored across languages (in the spirit of \href{https://www.cs.ru.nl/~freek/100/}{``Formalizing 100 Theorems"} and \href{https://www.themathgenome.com/}{The Math Genome Project})

\textbf{Scaling}. For those with more compute, scaling up the simulations (n\_generations, population\_size) will surely improve the confidence in the results and likely identify additional insights into longer evolutionary dynamics and performance. Furthermore, scaling in the training dimension by increasing max\_iters will enable more patient learning schedules and thus significantly impact evolution.

\textbf{Proof Checkers}. Integrating the respective proof checking environments to move beyond validation loss as a fitness metric (to rather use close \%) would be wonderful. 

\textbf{Where's Metamath?} Metamath's set-theoretic and single substitution tactic paradigm leads to a de Bruijn factor an order of magnitude larger \cite{GPT-f2020} than these type-theoretic and tactic-rich languages. The time and attention to properly integrate this into EvoGPT-f is left for future work.

\bibliographystyle{IEEEtran}
\bibliography{references}

\appendix
\section*{Appendix}

\section{Example to Compare Formal Math Languages}
\label{appendix:a}
\begin{figure}[H]
\centering
\includegraphics[width=4in]{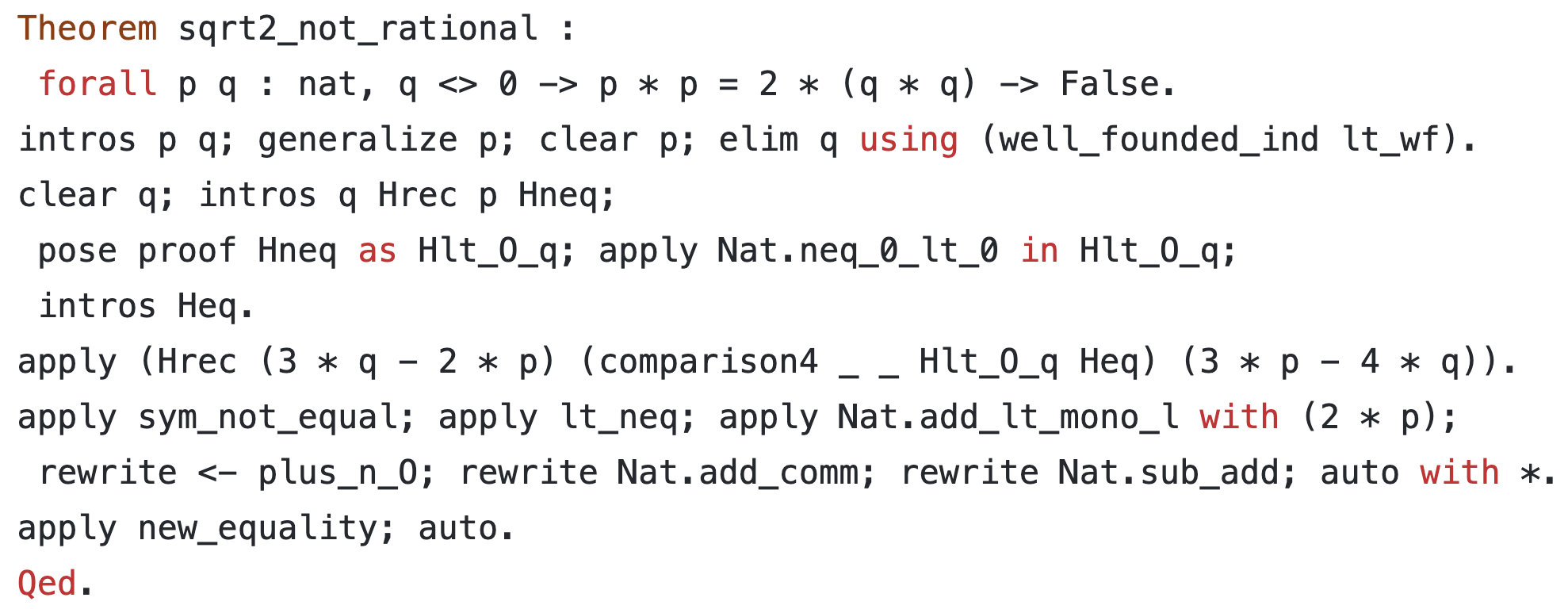}
\vspace{5mm}
\caption{\label{fig:coq_irrsqr2_figure}Coq theorem and proof of the irrationality of the square root of 2.}
\end{figure}

Coq is managed by the French National Institute for Research in Computer Science and Control (Inria). In Fig.~\ref{fig:coq_irrsqr2_figure}, a Coq proof of the irrationality of the square root of 2 is shown. 

\begin{figure}[H]
\centering
\includegraphics[width=4in]{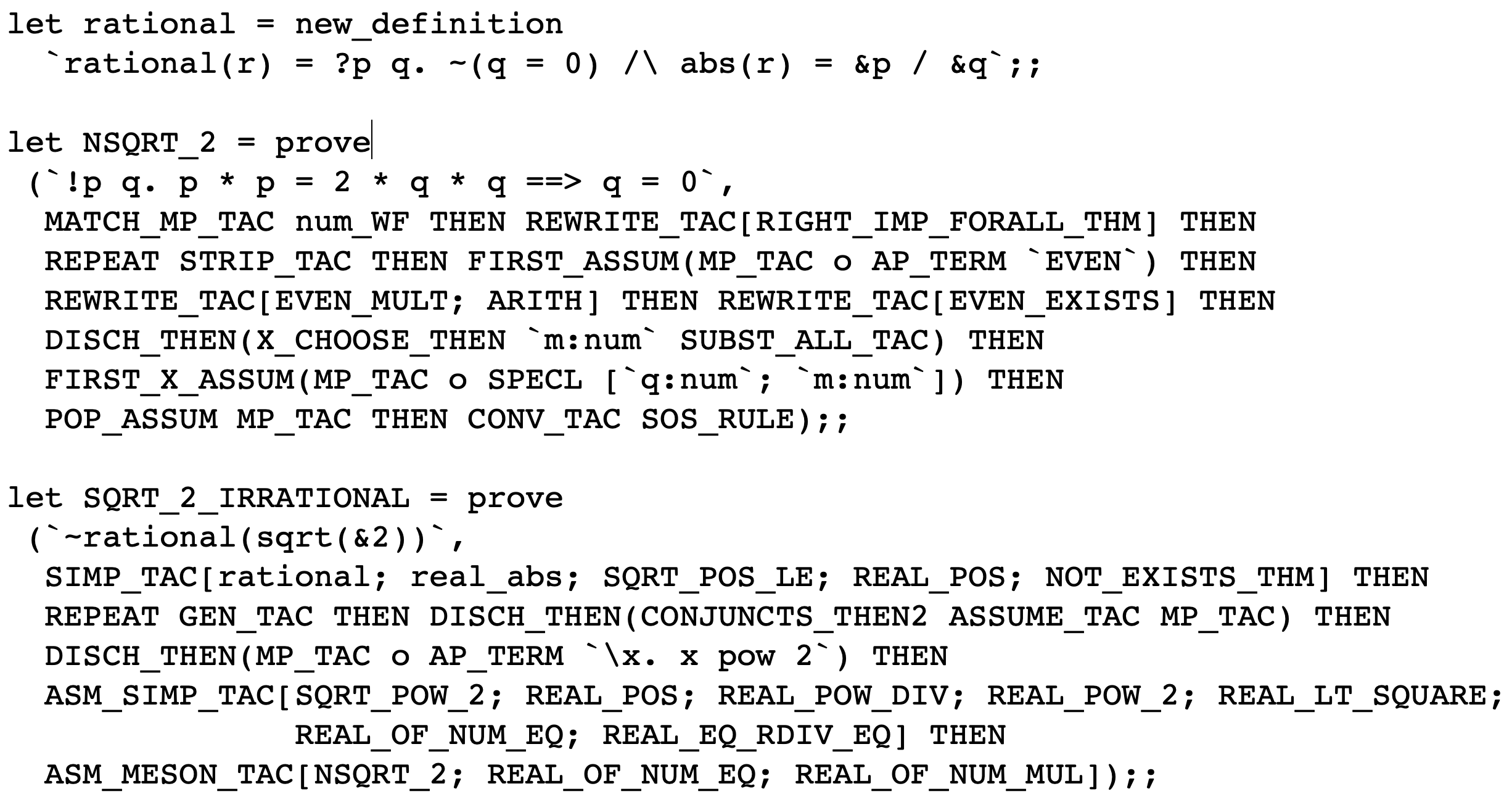}
\vspace{5mm}
\caption{\label{fig:hol_irrsqr2_figure} HOL Light theorem and proof of the irrationality of the square root of 2.}
\end{figure}

The HOL System was pioneered by Mike Gordon in the early 80's, and is developed and maintained by the University of Cambridge Computer Laboratory - Automated Reasoning Group, amongst others.\footnote{University of Cambridge Computer Laboratory - Automated Reasoning https://www.cl.cam.ac.uk/research/hvg/HOL/} See \cite{HOLHistory} for a history of HOL system development. In Fig.~\ref{fig:hol_irrsqr2_figure}, a HOL Light proof of the irrationality of the square root of 2 is shown. 

\begin{figure}[H]
\centering
\includegraphics[width=4in]{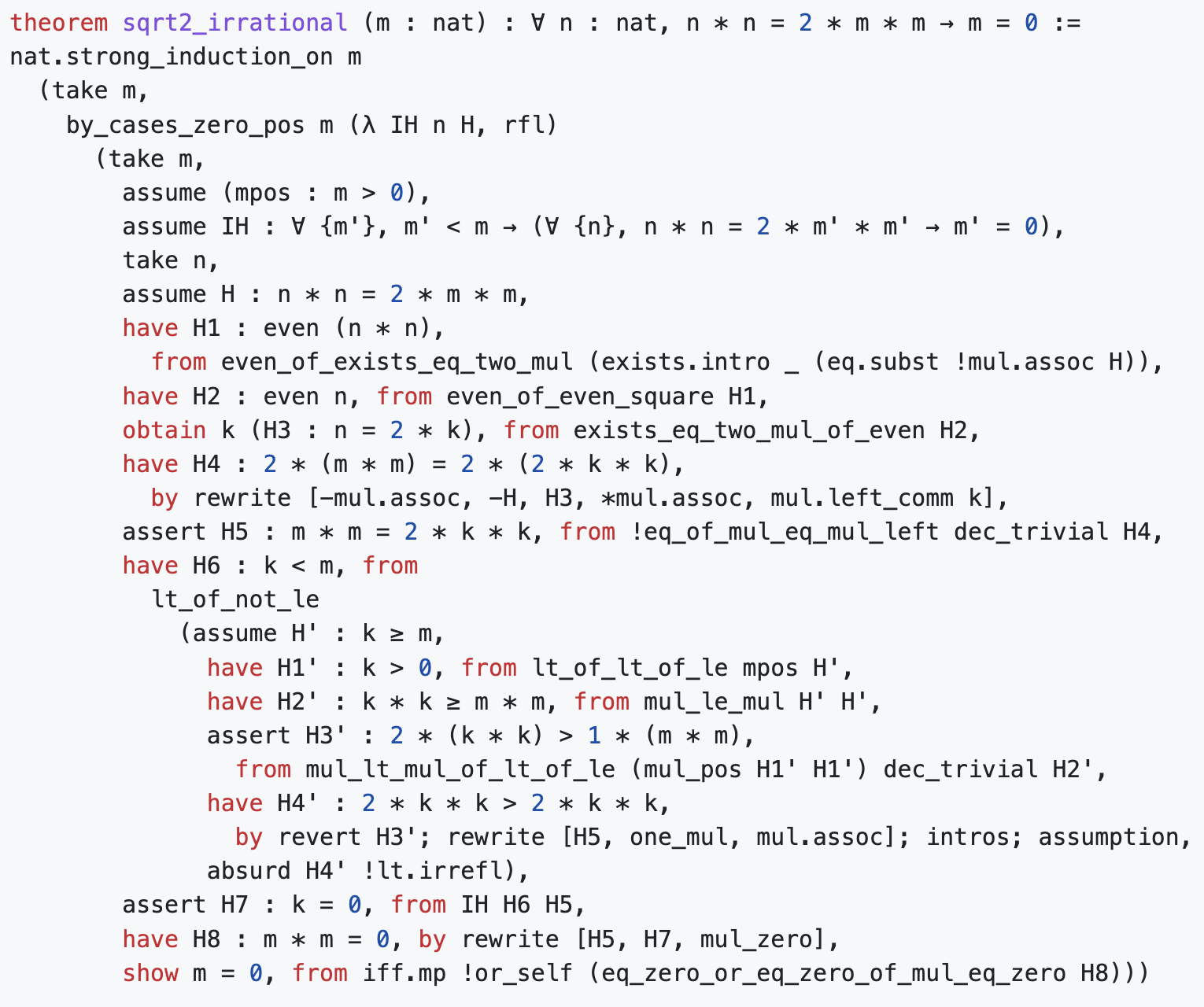}
\vspace{5mm}
\caption{Lean theorem and proof of the irrationality of the square root of 2}
\label{fig:lean_sqr2_proof}
\end{figure}

Developed by Leonardo de Moura at Microsoft Research, Lean is designed to be highly modular and extensible, with an emphasis on automation and metaprogramming capabilities. In Fig.~\ref{fig:lean_sqr2_proof}, a Lean 3 proof of the irrationality of the square root of 2 is shown.

These screenshots in Fig. ~\ref{fig:coq_irrsqr2_figure} - ~\ref{fig:lean_sqr2_proof} are from the code repositories provided in Table ~\ref{tab:table1}.

\vspace{5mm}

\vspace{5mm}

\newpage

\section{Byte Pair Encoding Vocabulary Size Selection}
\label{appendix:b}
\begin{figure}[H]
    \centering
    \includegraphics[width=1\textwidth]{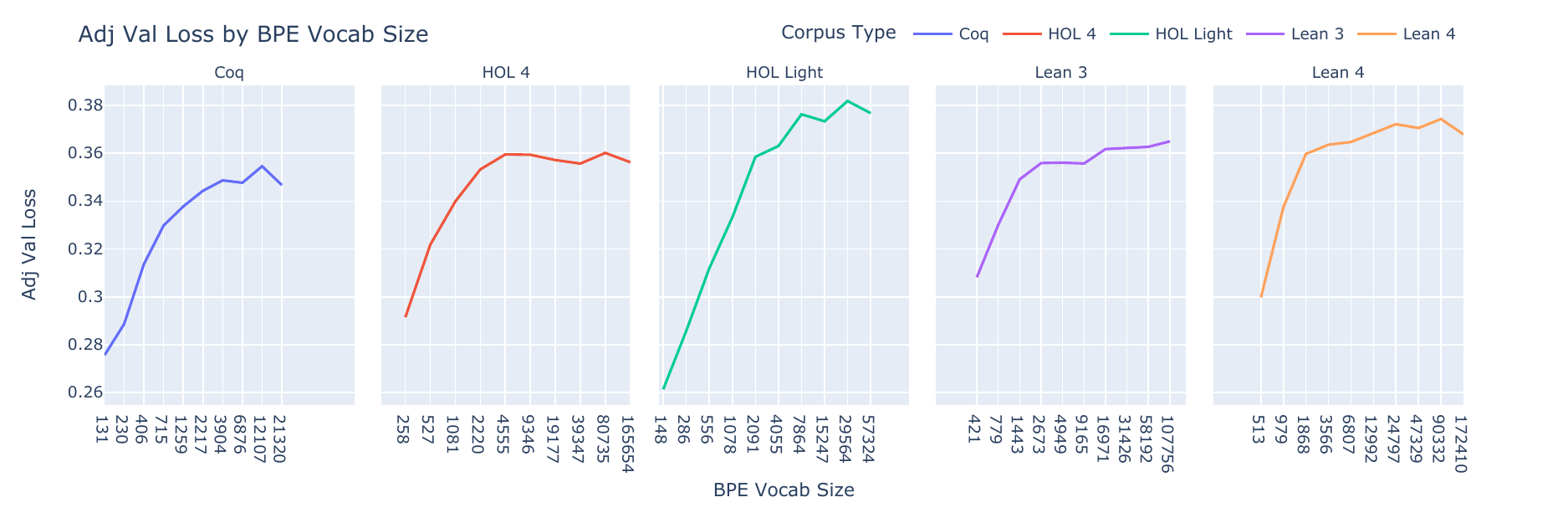}
    \caption{BPE Vocabulary Size Selection by Corpus}
    \label{fig:adj_val_loss_by_bpe_vocab_size_cf_corpus}
\end{figure}

\newpage
\section{Additional Distribution Details}

\subsection{Underlying Distributions of the Mean Loss Curves shown in Fig. ~\ref{fig:line_plot3_entropy_and_discrepancy_adj_val_loss_by_generation_cf_tkn_main}}
\label{appendix:c1}
\begin{figure}[H]
    \centering 
    \includegraphics[width=0.8\textwidth]{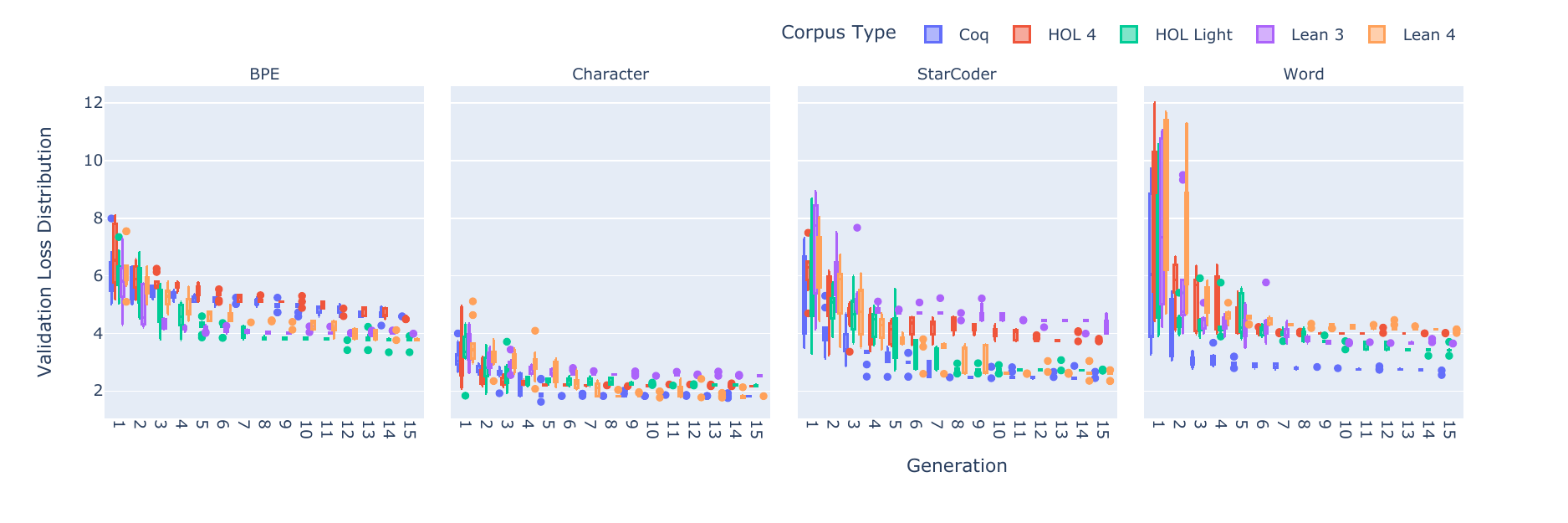}
    \label{fig:box_plot0_best_val_loss_by_generation_cf_tkn}
    \vspace{1mm}
    \includegraphics[width=0.8\textwidth]{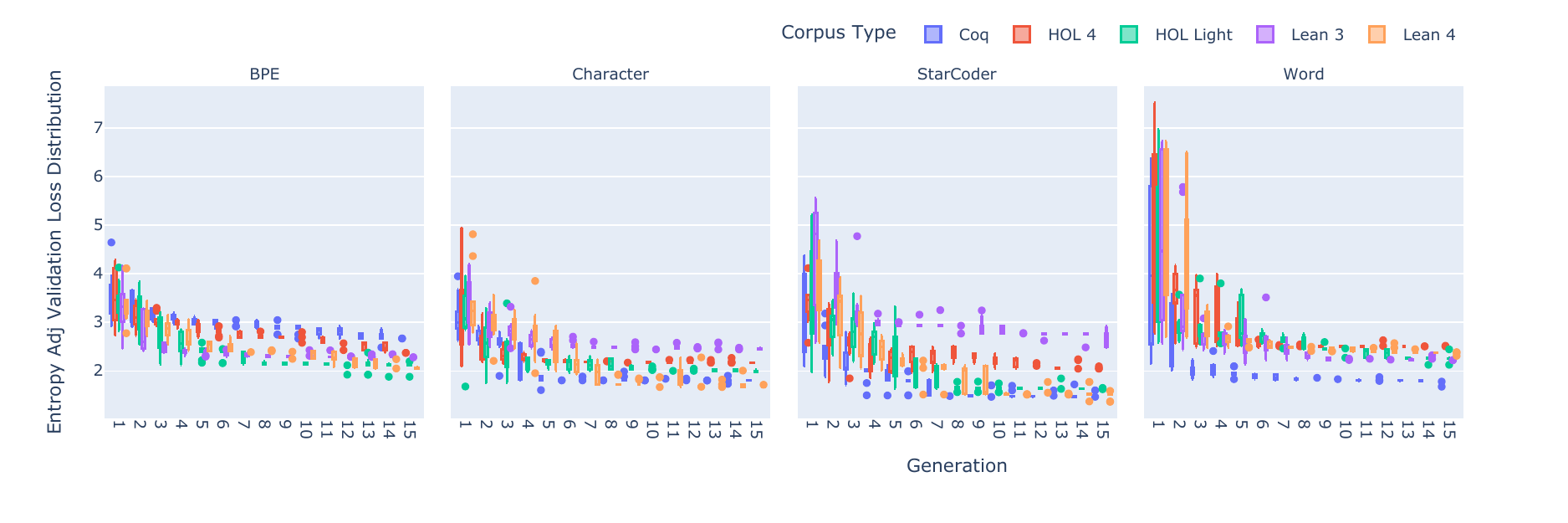}
    \label{fig:box_plot1_entropy_adj_val_loss_by_generation_cf_tkn}
    \vspace{1mm}

    \includegraphics[width=0.8\textwidth]{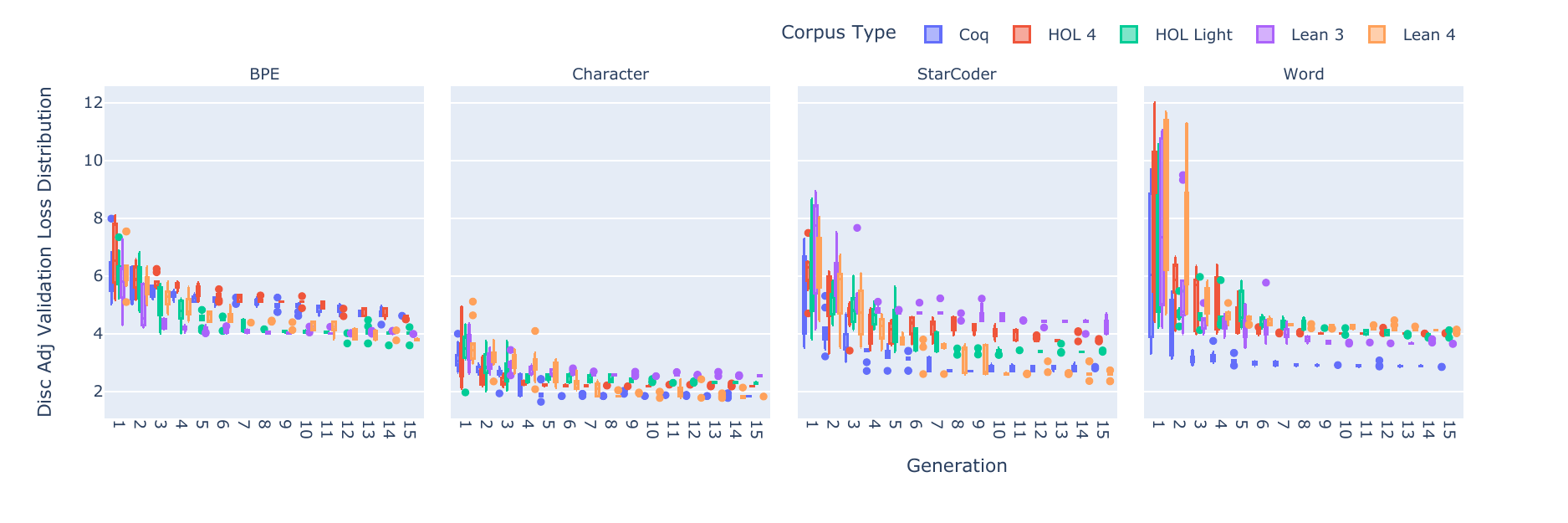}
    \label{fig:box_plot2_discrepancy_adj_val_loss_by_generation_cf_tkn}
    \vspace{1mm}

    \includegraphics[width=0.8\textwidth]{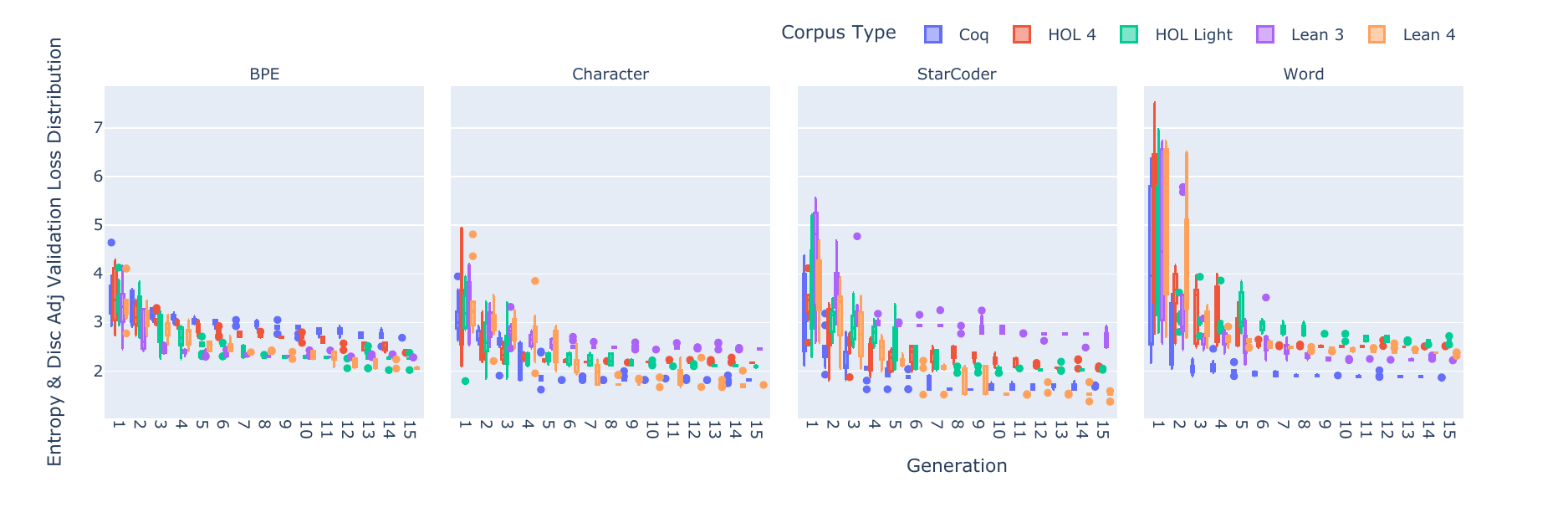}
    \caption{a.) Original Validation Loss, b.) Entropy Adjusted Validation Loss, c.) Overfitting Penalty Adjusted Validation Loss, d.) Final Adjusted Validation Loss}
    \label{fig:box_plot3_entropy_and_discrepancy_adj_val_loss_by_generation_cf_tkn}
\end{figure}
\newpage

\subsection{Underlying Distributions of the Mean Loss Curves by Corpus}
\label{appendix:c2}
This is a complementary view of loss distributions, by corpus type, rather than by tokenization method. Here, an additional visual encoding is included to encode the (smallest)  \tikz[baseline=-0.5ex]{
    \filldraw[fill={rgb,255:red,253; green,202; blue,38}] (0,0) rectangle (1ex,1ex);
    \node[anchor=west] at (1ex,0) {};
    \filldraw[fill={rgb,255:red,237; green,121; blue,83}] (3ex,0) rectangle (4ex,1ex);
    \node[anchor=west] at (4ex,0) {};
    \filldraw[fill={rgb,255:red,189; green,55; blue,134}] (6ex,0) rectangle (7ex,1ex);
    \node[anchor=west] at (7ex,0) {};
    \filldraw[fill={rgb,255:red,114; green,1; blue,168}] (9ex,0) rectangle (10ex,1ex);
    \node[anchor=west] at (10ex,0) {};
}(largest) vocabulary sizes of the respective tokenization methods. 

\begin{figure}[H]
    \centering 
    \includegraphics[width=0.8\textwidth]{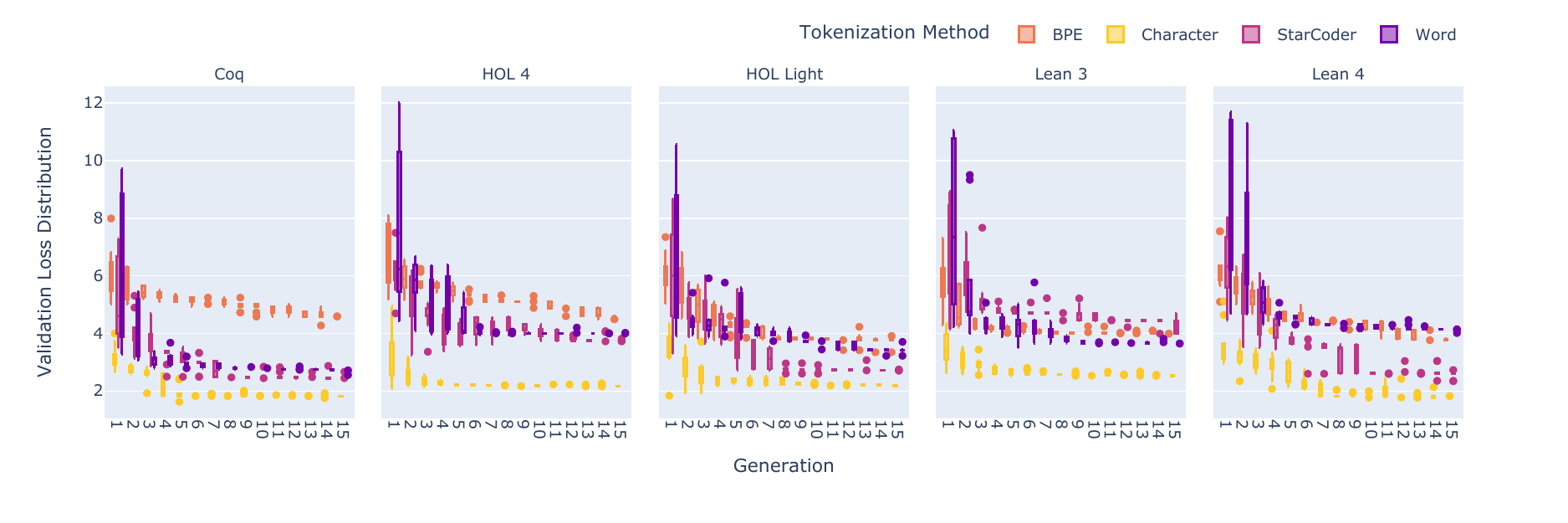}
    \label{fig:box_plot0_best_val_loss_cf_corpus}
    \vspace{1mm}

    \includegraphics[width=0.8\textwidth]{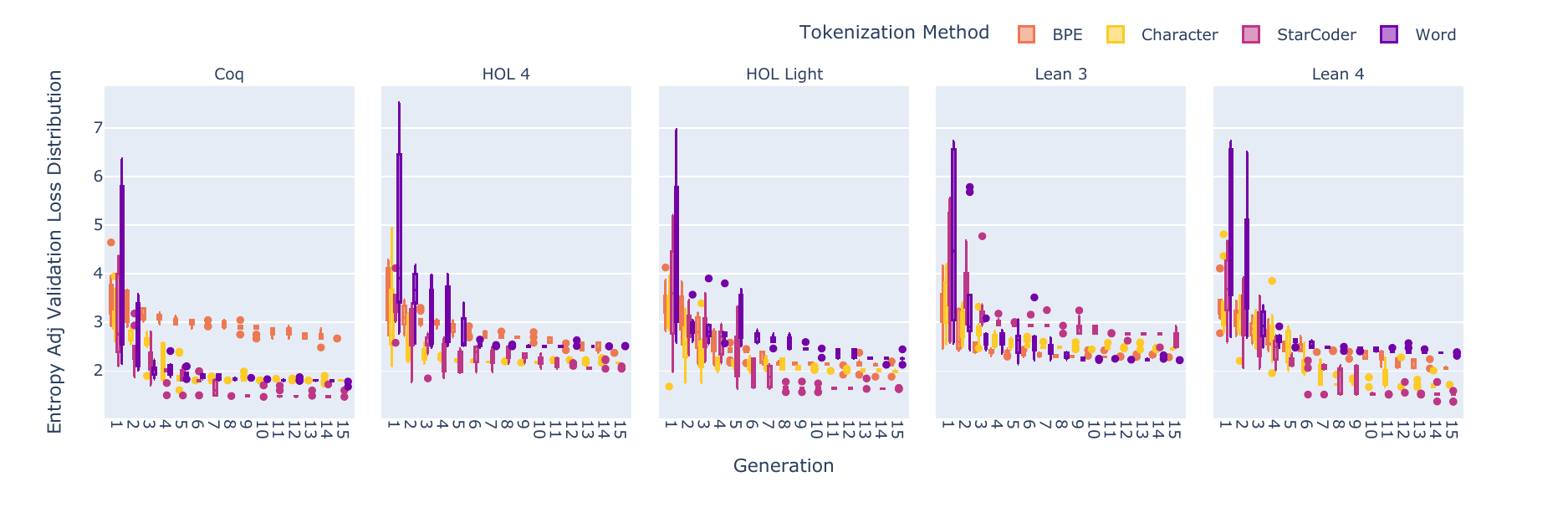}
    \label{fig:box_plot1_entropy_adj_val_loss_cf_corpus}
    \vspace{1mm}

    \includegraphics[width=0.8\textwidth]{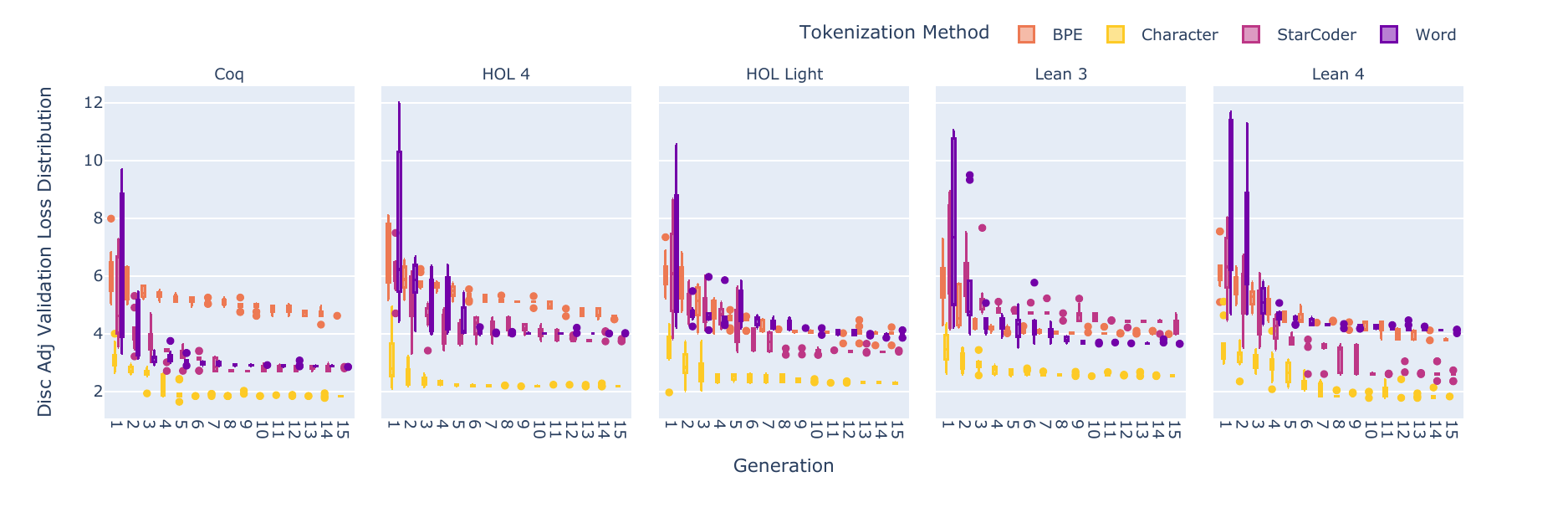}
    \label{fig:box_plot2_discrepancy_adj_val_loss_cf_corpus}
    \vspace{1mm}

    \includegraphics[width=0.8\textwidth]{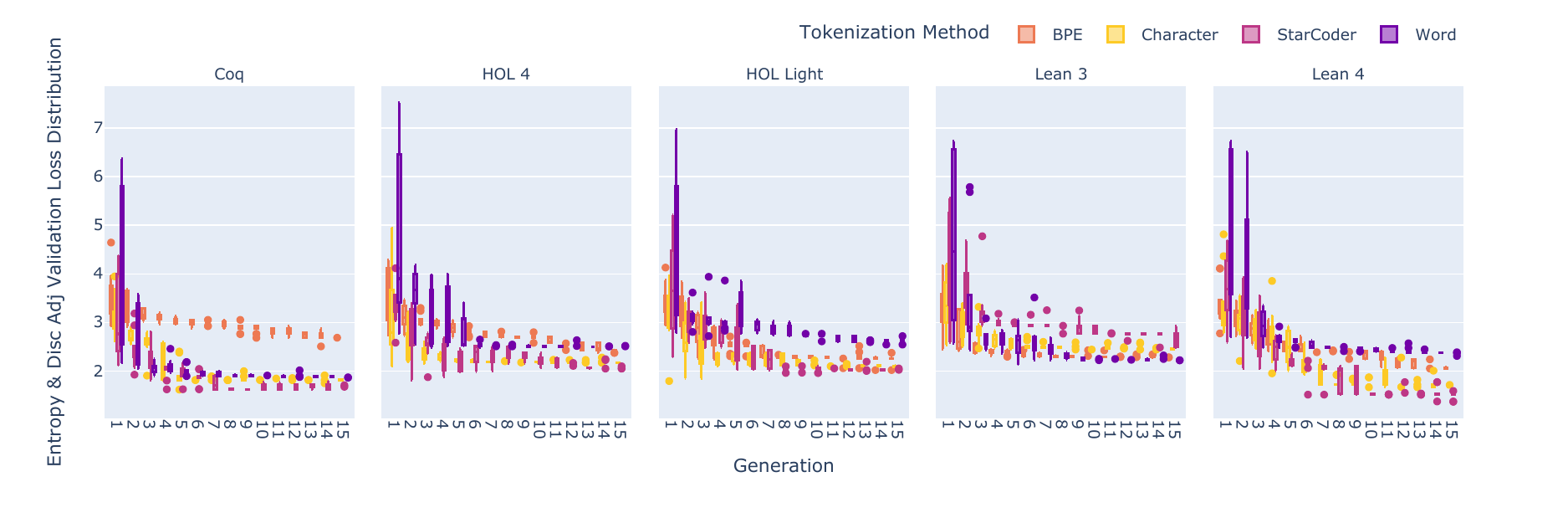}
    \caption{(Complementary View) Validation Loss Figures Grouped by Corpus}
    \label{fig:box_plot3_entropy_and_discrepancy_adj_val_loss_cf_corpus}
\end{figure}

\clearpage

\subsection{Modeling Details}
\label{appendix:c4}

\begin{figure}[H]
    \centering 
    \includegraphics[width=0.5\textwidth]{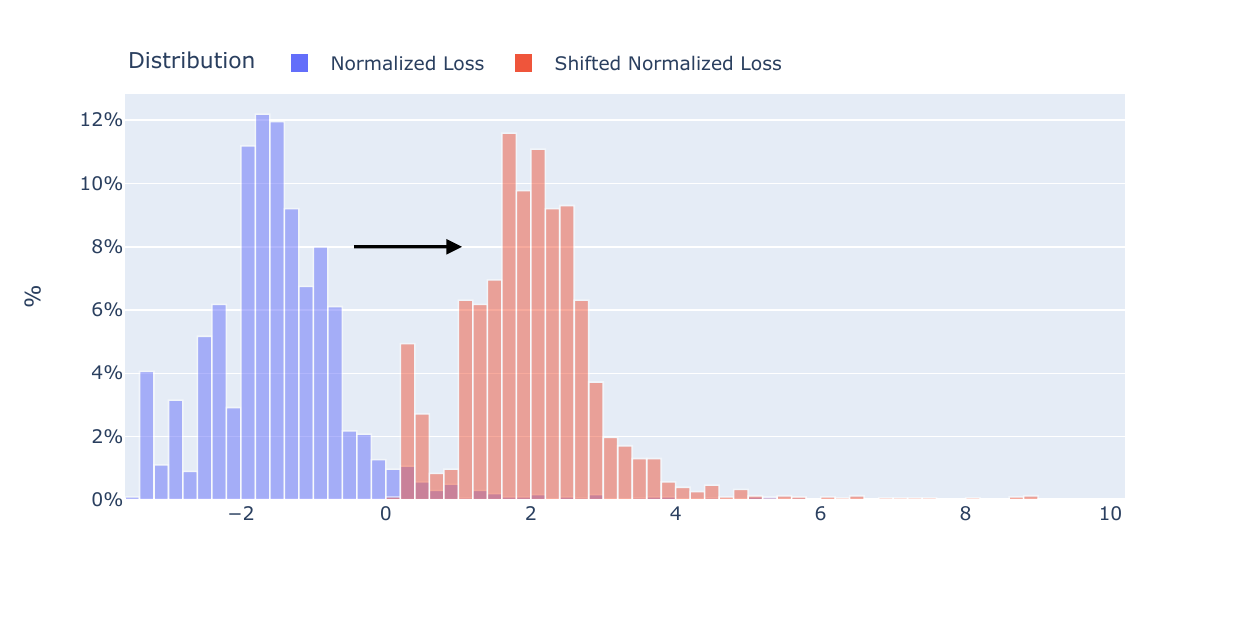}
    \caption{Original and Shifted Validation Loss}
    \label{fig:histograms_for_regression_models}
\end{figure}

\begin{table}[H]
\label{table:model_factors}
\begin{tabular}{lllrrl}
\toprule
                  Response &           Model &    Feature &  Factor &  P-Value & Factor 95\% C.I. \\
\midrule
  Shifted Information Gain & Wtd GLM \textasciitilde  Gamma &  Intercept &    1.37 &     0.00 &    (1.21, 1.57) \\
                           &                 &        Coq &    1.00 &     0.96 &     (0.83, 1.2) \\
                           &                 &      HOL 4 &    1.28 &     0.01 &    (1.07, 1.54) \\
                           &                 &     Lean 3 &    1.46 &     0.00 &    (1.22, 1.76) \\
                           &                 &     Lean 4 &    0.85 &     0.09 &    (0.71, 1.02) \\
                           &     GEE \textasciitilde  Gamma &  Intercept &    3.54 &     0.00 &    (2.88, 4.34) \\
                           &                 &        Coq &    0.88 &     0.00 &    (0.82, 0.94) \\
                           &                 &      HOL 4 &    1.10 &     0.01 &    (1.03, 1.18) \\
                           &                 &     Lean 3 &    1.00 &     0.85 &    (0.97, 1.04) \\
                           &                 &     Lean 4 &    0.92 &     0.00 &     (0.9, 0.95) \\
                           &                 & Generation &    0.93 &     0.00 &    (0.92, 0.95) \\
                           & Wtd GEE \textasciitilde  Gamma &  Intercept &    2.93 &     0.00 &      (2.6, 3.3) \\
                           &                 &        Coq &    0.95 &     0.00 &    (0.95, 0.95) \\
                           &                 &      HOL 4 &    1.19 &     0.00 &    (1.19, 1.19) \\
                           &                 &     Lean 3 &    1.28 &     0.00 &    (1.28, 1.29) \\
                           &                 &     Lean 4 &    0.97 &     0.00 &    (0.97, 0.97) \\
                           &                 & Generation &    0.94 &     0.00 &    (0.93, 0.96) \\
Calibrated Validation Loss & Wtd GLM \textasciitilde  Gamma &  Intercept &    2.25 &     0.00 &    (2.15, 2.35) \\
                           &                 &        Coq &    0.90 &     0.00 &    (0.85, 0.96) \\
                           &                 &      HOL 4 &    1.02 &     0.51 &    (0.96, 1.08) \\
                           &                 &     Lean 3 &    1.08 &     0.02 &    (1.01, 1.14) \\
                           &                 &     Lean 4 &    0.85 &     0.00 &      (0.8, 0.9) \\
                           &     GEE \textasciitilde  Gamma &  Intercept &    3.56 &     0.00 &    (3.16, 4.01) \\
                           &                 &        Coq &    0.85 &     0.00 &    (0.82, 0.88) \\
                           &                 &      HOL 4 &    0.96 &     0.02 &    (0.93, 0.99) \\
                           &                 &     Lean 3 &    0.97 &     0.00 &    (0.95, 0.99) \\
                           &                 &     Lean 4 &    0.95 &     0.00 &    (0.94, 0.95) \\
                           &                 & Generation &    0.96 &     0.00 &    (0.95, 0.97) \\
                           & Wtd GEE \textasciitilde  Gamma &  Intercept &    3.20 &     0.00 &    (3.03, 3.37) \\
                           &                 &        Coq &    0.89 &     0.00 &    (0.89, 0.89) \\
                           &                 &      HOL 4 &    1.01 &     0.00 &    (1.01, 1.02) \\
                           &                 &     Lean 3 &    1.05 &     0.00 &    (1.05, 1.05) \\
                           &                 &     Lean 4 &    0.93 &     0.00 &    (0.93, 0.93) \\
                           &                 & Generation &    0.97 &     0.00 &    (0.96, 0.98) \\
\bottomrule
\end{tabular}
\vspace{1mm}
\caption{Inferences about all Model Parameters. Factors are defined as $e^{\beta_{i}}$. Corpus factors are \\\hspace{\textwidth} interpreted as the $100*(e^{\beta_{i}}-1)$ percent change in the response from the baseline corpus (HOL Light).}
\end{table}
\newpage

\subsection{EvoEDA - Application for Exploratory Analysis}

\label{appendix:c5}
\begin{figure}[H]
    \centering 
    \includegraphics[width=0.6\textwidth]{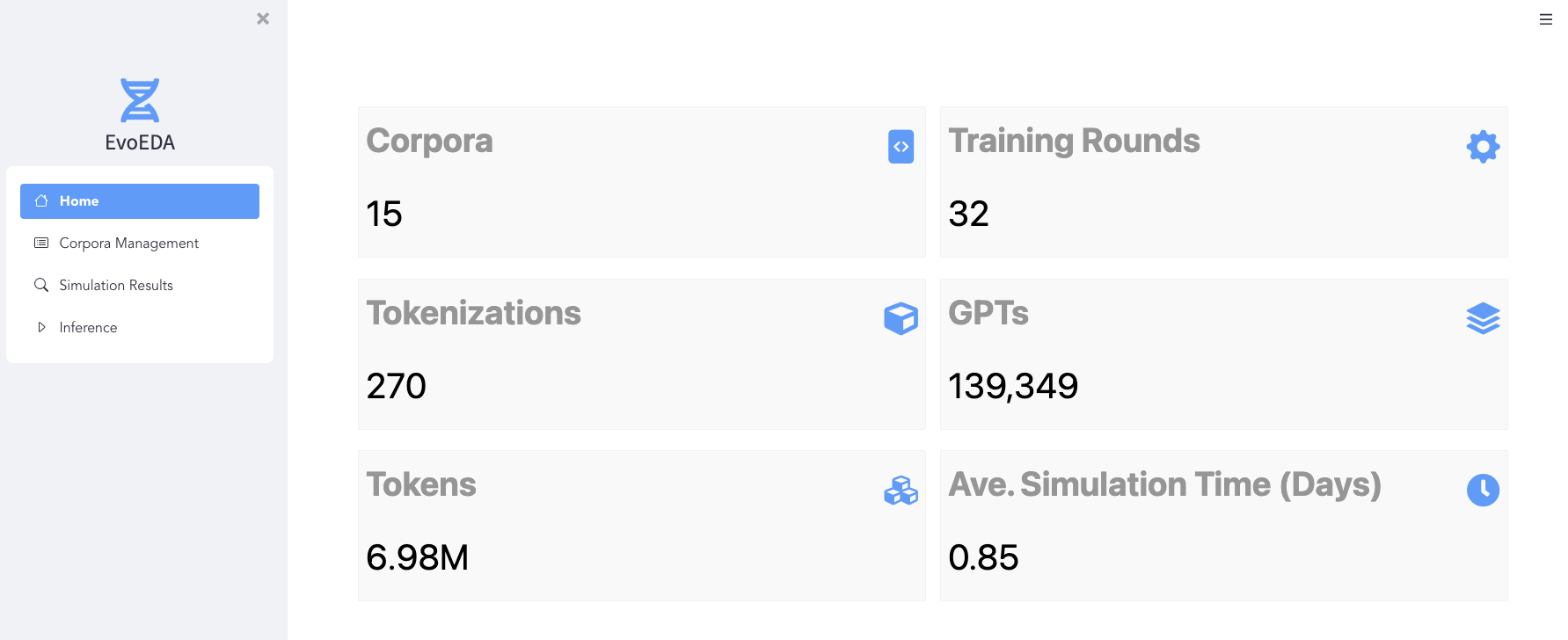}
    \vspace{1mm}
    \caption{EvoEDA - Statistics Dashboard}
    \label{fig:evo_eda_app0}
\end{figure}
\vspace{2mm}
\begin{figure}[H]
    \centering 
    \includegraphics[width=0.75\textwidth]{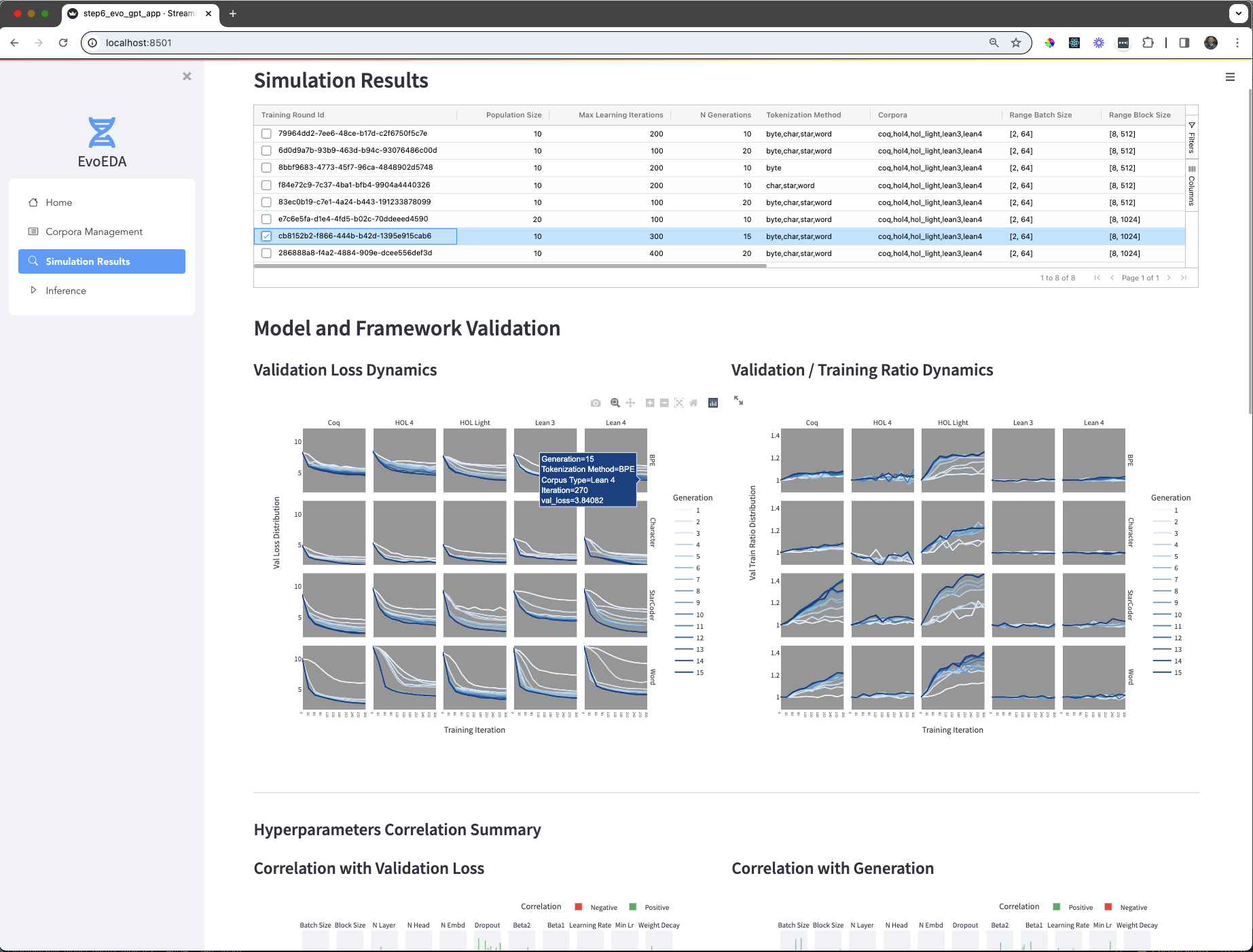}
    \label{fig:evo_eda_app1}
    \vspace{1mm}
    \caption{EvoEDA - Dynamic Training Round Exploratory Analysis}
\end{figure}
\vspace{2mm}
\begin{figure}[H]
    \centering 
    \includegraphics[width=0.8\textwidth]{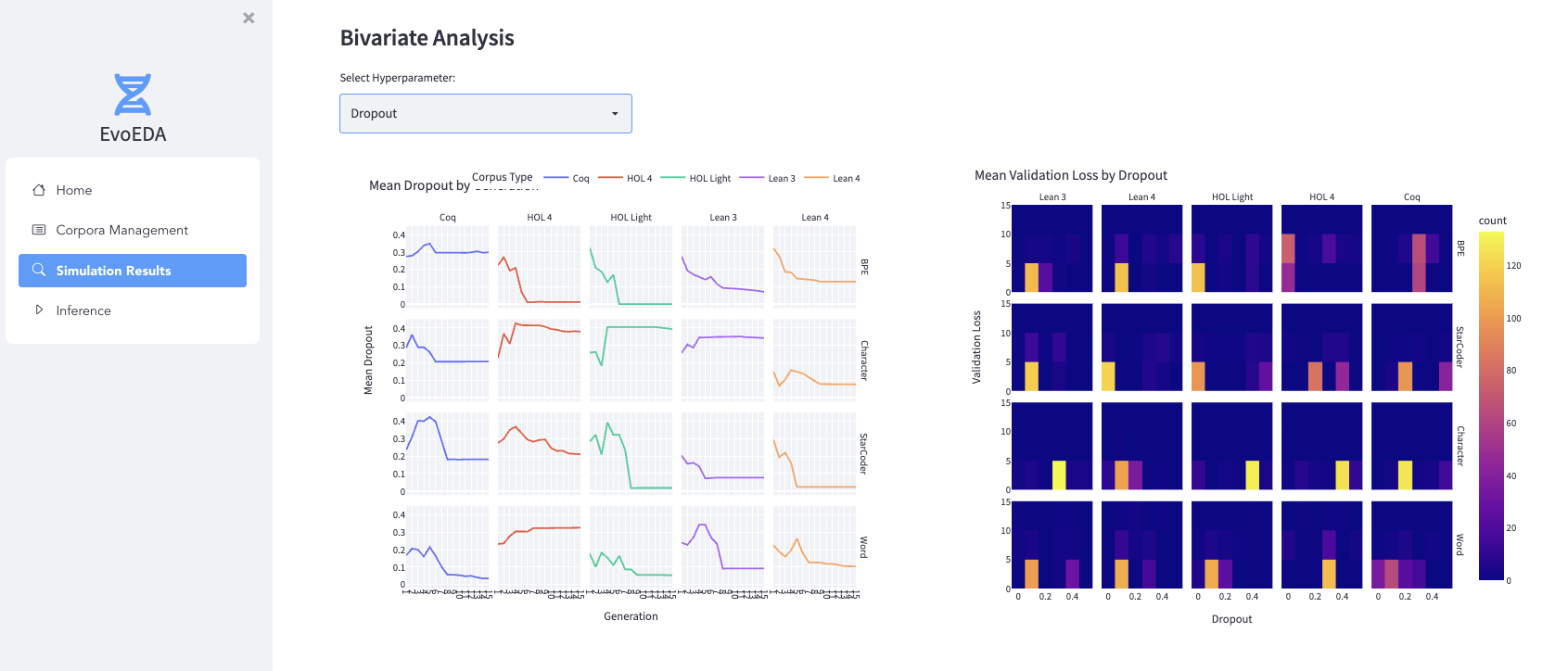}
    \label{fig:evo_eda_app2}
    \vspace{1mm}
    \caption{EvoEDA - Drill-down Bivariate Analysis}
\end{figure}

\end{document}